\documentclass{article} 
\pdfoutput=1
\usepackage{nips12submit_e,times}

\usepackage{graphicx} 
\usepackage{subfigure} 
\usepackage{tabularx}
\usepackage{natbib}
\usepackage{verbatim}
\usepackage{wrapfig}

\usepackage{my_gp_macros}
\usepackage{amsmath}
\usepackage{amssymb}
\usepackage{amsfonts}
\usepackage{parskip}
\usepackage{paralist}
\newcommand{\wt}[1]{\widetilde{#1}}
\newcommand{\ty}{\wt{y}}
\newcommand{\tL}{\wt{L}}
\renewcommand{\comment}[2]{#2}

\title{Efficient Optimization for \\
  Sparse Gaussian Process Regression}
 
\author{
	Yanshuai Cao${}^1$ ~~
	Marcus A.~Brubaker${}^2$ ~~
	David J.~Fleet${}^1$ ~~ 
	Aaron Hertzmann${}^{1,3}$ \\[2ex]
	\begin{tabular}{ccc}
	${}^1$Department of Computer Science &
	${}^2$TTI-Chicago &
	${}^3$Adobe Research \\
	University of Toronto
	\end{tabular}
}

\nipsfinalcopy 

\begin{document}
\maketitle

\begin{abstract}
\vspace*{-0.2cm}
\comment{
We propose an efficient optimization algorithm for selecting a subset 
of training data for sparse Gaussian process regression. The algorithm 
estimates a sparsity inducing set and the hyperparameters using a 
single objective, formulated in terms of either marginal likelihood or 
variational bound.  
The space and time complexity are linear in the training set 
size, and the algorithm can be applied to large regression problems on
discrete or continuous domains. Empirical evaluation shows
state-of-art performance in the discrete case and competitive
results in the continuous case.}

We propose an efficient optimization algorithm for selecting a subset 
of training data to induce sparsity for Gaussian process regression. 
The algorithm estimates an inducing set and the hyperparameters using 
a single objective, either the marginal likelihood or a variational 
free energy.  
The space and time complexity are linear in training set size, and 
the algorithm can be applied to large regression problems on
discrete or continuous domains. Empirical evaluation shows
state-of-art performance in discrete cases and competitive
results in the continuous case.
\end{abstract}

\vspace*{-0.2cm}

\section{Introduction}

\vspace*{-0.15cm}

Gaussian Process (GP) learning and inference are computationally 
prohibitive with large datasets, having time complexities $O(n^3)$ 
and $O(n^2)$, where $n$ is the number of training points.
Sparsification algorithms exist that scale linearly in the training 
set size (see \cite{Quinonero2005} for a review). They construct a 
low-rank approximation to the GP covariance matrix over the full 
dataset using a small set of {\em inducing points}. 
Some approaches select inducing points from training points
\cite{Keerthi06amatching,Lawrence03,Seeger03,Smola01sparsegreedy}.
But these methods select the inducing points using ad hoc criteria; 
i.e., they use different objective functions to select inducing points 
and to optimize GP hyperparameters.
More powerful sparsification methods \cite{Snelson06,Titsias09,Walder2008} 
use a single objective function and allow inducing points to move
freely over the input domain which are learned via gradient descent.
This continuous relaxation is not feasible, however, if the input 
domain is discrete, or if the kernel function is not differentiable 
in the input variables.
As a result, there are problems in myraid domains, like bio-informatics,
linguistics and computer vision where current sparse GP regression 
methods are inapplicable or ineffective.

We introduce an efficient sparsification algorithm for GP regression.  
The method optimizes a single objective for joint selection of 
inducing points and GP hyperparameters.  
Notably, it optimizes either the marginal likelihood, or a variational 
free energy \cite{Titsias09}, exploiting the QR factorization of a partial Cholesky 
decomposition to efficiently approximate the covariance matrix. 
Because it chooses inducing points from the training data, it is
applicable to problems on discrete or continuous input domains. 
To our knowledge, it is the first method for selecting discrete inducing 
points and hyperparameters that optimizes a single objective, with 
linear space and time complexity. 
It is shown to outperform other methods on discrete datasets from bio-informatics and computer vision.  
On continuous domains it is competitive with the Pseudo-point 
GP \cite{Snelson06} (SPGP).

\vspace*{-0.15cm}
\subsection{Previous Work}
\vspace*{-0.15cm}

Efficient state-of-the-art sparsification methods are $O(m^2 n)$ 
in time and $O(mn)$ in space for learning.  They compute the 
predictive mean and variance in time $O(m)$ and $O(m^2)$.
Methods based on continuous relaxation, when applicable, entail
learning $O(md)$ continuous parameters, where $d$ is the input dimension.
In the discrete case, combinatorial optimization is required to select
the inducing points, and this is, in general, intractable.
Existing discrete sparsification methods therefore use other criteria 
to greedily select inducing points
\cite{Keerthi06amatching,Lawrence03,Seeger03,Smola01sparsegreedy}.
Although their criteria are justified, each in their own way
({\em e.g.}, \cite{Lawrence03,Seeger03} take an information theoretic
perspective),
they are greedy and do not use the same objective to select inducing 
points and to estimate GP hyperparameters.

The variational formulation of Titsias \cite{Titsias09} treats inducing 
points as variational parameters, and gives a unified objective for 
discrete and continuous inducing point models. In the 
continuous case, it uses gradient-based optimization to find inducing 
points and hyperparameters.
In the discrete case, our method optimizes the same variational
objective of Titsias \cite{Titsias09}, but is a significant improvement over greedy forward
selection using the variational objective as selection criteria, or
some other criteria. In particular, given the cost of evaluating the
variational objective on all training points, Titsias \cite{Titsias09} evaluates
the objective function on a small random subset of candidates at each
iteration, and then select the best element from the subset. This
approximation is often slow to achieve good results, as we explain and demonstrate below in Section \ref{subsec:discrete_domain_exp}. 
The approach in \cite{Titsias09} also uses greedy forward 
selection, which provides no way to refine the inducing set after hyperparameter 
optimization, except to discard all previous inducing points and 
restart selection. Hence, the objective is not guaranteed to 
decrease after each restart.
By comparison, our formulation considers all candidates at each step, 
and revisiting previous selections is efficient, and guaranteed to 
decrease the objective or terminate.

Our low-rank decomposition is inspired by the {\em Cholesky with Side 
Information} (CSI) algorithm for kernel machines \cite{Bach05}.  We extend 
that approach to GP regression.  First, we alter the form of the low-rank 
matrix factorization in CSI to be suitable for GP regression with full-rank 
diagonal term in the covariance.
Second, the CSI algorithm selects inducing points in a single greedy 
pass using an approximate objective.  We propose an iterative optimization 
algorithm that swaps previously selected points with new candidates 
that are guaranteed to lower the objective.  Finally, we perform inducing 
set selection jointly with gradient-based hyperparameter estimation instead 
of the grid search in CSI.
Our algorithm selects inducing points in a principled fashion, optimizing 
the variational free energy or the log likelihood.  It does so with time 
complexity $O(m^2n)$, and in practice provides an improved quality-speed 
trade-off over other discrete selection methods.

\vspace*{-0.1cm}

\section{Sparse GP Regression}

\vspace*{-0.2cm}

Let $y \in \mathbb{R}$ be the noisy output of a function, $f$,  
of input $\vc{x}$. 
Let $X = \{ \vc{x}_i \}_{i=1}^{n}$ denote $n$ training
inputs, each belonging to input space $\mathcal{D}$, which is not
necessarily Euclidean.
Let $\vc{y} \in \mathbb{R}^n$ denote the corresponding vector of 
training outputs.  Under a full zero-mean GP, with the covariance function
\begin{equation}
\mE [ y_i y_j ] ~=~ \kappa(\vc{x}_i, \vc{x}_j) + \sigtwo 1[i=j] ~,
\end{equation}
where $\kappa$ is the kernel function, $1[\cdot]$ is the usual 
indicator function, and $\sigma^2$ is the variance of the observation
noise, the predictive distribution over the output $f_{\star}$ at a 
test point $\vc{x}_{\star}$ is normally distributed.  The mean and variance
of the predictive distribution can be expressed as 
\begin{align*}
\mu_{\star} & ~=~
\trans{\vc{\kappa}(\vc{x}_{\star})} \left(K + \sigtwo I_{n}\right)^{-1}\vc{y}\\
v^2_{\star} & ~=~
\kappa(\vc{x}_{\star},\vc{x}_{\star}) - 
\trans{\vc{\kappa}(\vc{x}_{\star})} 
\left(K + \sigtwo I_{n}\right)^{-1} 
\vc{\kappa}(\vc{x}_{\star})
\end{align*}
where $I_n$ is the $n\times n$ identity matrix, $K$ is the kernel matrix 
whose $ij$th element is $\kappa(\vc{x}_i, \vc{x}_j)$, and 
$\vc{\kappa}(\vc{x}_{\star})$ is the column vector whose $i$th element 
is $\kappa(\vc{x}_{\star}, \vc{x}_i)$.

The hyperparameters of a GP, denoted $\vc{\theta}$, comprise the 
parameters of the kernel function, and the noise variance $\sigtwo$.  
The natural objective for learning  $\vc{\theta}$ is the negative 
marginal log likelihood (NMLL) of the training data,  
$-\log \left(P(\vc{y}|X,\vc{\theta})\right) $, given up to 
a constant by
\begin{equation} 
\label{eq:full_gp_nmll}
E_{\mathit{full}}(\vc{\theta}) ~= ~ \, 
(\, \vr{y}\!\left(K \!+\!  \sigtwo I_{n} \right)^{-1}\! \vc{y} 
+ 
\log | K \!+\! \sigtwo I_{n} | \, )\,  /\, 2 ~.
\end{equation}
The computational bottleneck lies in the $O(n^2)$ storage and $O(n^3)$
inversion of the full covariance matrix, $K + \sigtwo I_n$.  To lower 
this cost with a sparse approximation, Csat\'{o} and Opper \cite{Csato2002} 
and Seeger {\it et al.} \cite{Seeger03} proposed the Projected Process (PP) 
model, wherein a set of $m$ inducing points are used to construct a low-rank 
approximation of the kernel matrix.
In the discrete case, where the inducing points are a subset of the 
training data, with indices $\mI \subset \{1,2,...,n\}$,
this approach amounts to replacing the kernel matrix $K$ with the 
following Nystr\"{o}m approximation \cite{rasmussen2006gaussian}:
\begin{equation} 
\label{eq:nystrom_approx}
K\,\simeq\, \hat{K} \,=\, K[:,\mI]\, K[\mI,\mI]^{-1}\, K[\mI,:]
\end{equation}
where $K[:,\mI]$ denotes the sub-matrix of $K$ comprising columns 
indexed by $\mI$, and $K[\mI,\mI]$ is the sub-matrix of 
$K$ comprising rows and columns indexed by $\mI$. We assume the rank 
of $K$ is $m$ or higher so we can always find such rank-$m$ approximations.
The PP NMLL is then algebraically equivalent to replacing $K$ with 
$\hat{K}$ in Eq.\,(\ref{eq:full_gp_nmll}), {\em i.e.}, 
\begin{equation} 
\label{eq:pp_gp_nmll}
E(\vc{\theta},\mI) ~=~ 
\left( \, 
E^{D}(\vc{\theta},\mI) \, + \, E^{C}(\vc{\theta},\mI) 
\, \right) / 2 ~,
\end{equation}
with data term 
$E^{D}(\vc{\theta},\mI)=\vr{y}( \!\hat{K}\! +\!\sigtwo I_{n})^{\! -1}\vc{y}$,
and model complexity 
$E^{C}(\vc{\theta},\mI) =  \log|\hat{K} \! +\! \sigtwo I_{n}|$.

The computational cost reduction from $O(n^3)$ to $O(m^2n)$ associated 
with the new likelihood is achieved by applying the Woodbury inversion 
identity to $E^{D}(\vc{\theta},\mI)$ and $E^{C}(\vc{\theta},\mI)$.  
\comment{The resulting sparse GP is called the projected latent variable 
GP or the Projected Process (PP) GP.}  
The objective in (\ref{eq:pp_gp_nmll}) can be viewed as an approximate 
log likelihood for the full GP model, or as the exact log likelihood 
for an approximate model, called the Deterministically Trained 
Conditional \cite{Quinonero2005}.  

The same PP model can also be obtained by a variational argument, as
in \cite{Titsias09}, for which the variational free energy objective 
can be shown to be Eq.\,(\ref{eq:pp_gp_nmll}) plus one extra term; 
{\em i.e.},
\begin{equation} \label{eq:F}
 F(\vc{\theta},\mI)  ~=~ 
\left( \, 
E^{D}\!(\vc{\theta},\mI) + 
E^{C}\!(\vc{\theta},\mI) + 
E^{V}\!(\vc{\theta},\mI) 
\, \right) \, / \, 2 ~,
\end{equation}
where $E^{V}(\vc{\theta},\mI) = \sigma^{-2}\tr(K \! - \! \hat{K})$
arises from the variational formulation.  It effectively regularizes 
the trace norm of the approximation residual of the covariance matrix. 
The kernel machine of \cite{Bach05} also uses a 
regularizer of the form $\lambda\tr(K \! - \! \hat{K})$, however
$\lambda$ is a free parameter that is set manually.

\vspace*{-0.1cm}
\section{Efficient optimization}
\vspace*{-0.15cm}

We now outline our algorithm for optimizing the variational free 
energy (\ref{eq:F}) to select the inducing set $\mI$ and the 
hyperparameters $\vc{\theta}$.
(The negative log-likelihood (\ref{eq:pp_gp_nmll}) is similarly
minimized by simply discarding the $E^V$ term.)
The algorithm is a form of hybrid coordinate descent that alternates 
between discrete optimization of inducing points, and continuous 
optimization of the hyperparameters. 
%
%
We first describe the algorithm to select inducing points, and then 
discuss continuous hyperparameter optimization and termination 
criteria in Sec.\ \ref{sec:Hyperparams}.



Finding the optimal inducing set is a combinatorial problem;
global optimization is intractable.
Instead, the inducing set is initialized to a 
random subset of the training data, which is then refined by 
a fixed number of swap updates at each iteration.%
\footnote{The inducing set can be incrementally constructed,
as in \cite{Bach05}, however we found no benefit to this.}
In a single swap update, a randomly chosen inducing point is 
considered for replacement. 
If swapping does not 
improve the objective, then the original point is retained. 
There are $n - m$ potential replacements for each each 
swap update; the key is to efficiently determine which will maximally 
improve the objective.
With the techniques described below, the computation time required
to approximately evaluate all possible candidates and swap an
inducing point is $O(m n)$.
Swapping all inducing points once takes $O(m^2n)$ time.

\vspace*{-0.15cm} 
\subsection{Factored representation}
\vspace*{-0.2cm}


To support efficient evaluation of the objective and swapping, we use
a factored representation of the kernel matrix.  Given an inducing 
set $\mI$ of $k $ points, for any $k \le m$, 
the low-rank Nystr\"{o}m approximation 
to the kernel matrix (Eq. \ref{eq:nystrom_approx}) can be expressed in terms of 
a partial Cholesky factorization:
\begin{equation} 
\label{eq:equivalence}
\hat{K} ~=~ K[:,\mI] \, K[\mI,\mI]^{-1}\, K[\mI,:] ~=~ L(\mI)L(\mI)^{\T} ~,
\end{equation}
where $L(\mI) \in \Rm{n}{k}$ is, up to permutation of rows, lower
trapezoidal matrix ({\em i.e.}, has a $k \times k$ top lower triangular
block, again up to row permutation).
\comment{(as explained in detail in the supplementary material)}
\comment{{
taking the ordered sequence $\mI$ as pivots, the standard partial Cholesky 
factorization with
column pivoting on $K$ (see supplementary material for detailed algorithm)
produces a permutation $\mP$ of $(1 \ldots n)$, and a matrix $L(\mI)
\in \Rm{n}{k}$. After permuting the rows of $L(\mI)$ with $\mP$, the
matrix $L[\mP,:]$ is lower triangular and is the partial Cholesky
factor of $K[\mP,\mP]$. Furthermore, the partial Cholesky
factorization and Nystr\"{o}m approximation \ref{eq:nystrom_approx} 
are equivalent:
\begin{equation} \label{eq:equivalence}
\hat{K} ~=~ K[:,\mI] \, K[\mI,\mI]^{-1}\, K[\mI,:] ~=~ L(\mI)L(\mI)^{\T} ~.
\end{equation}
}}
The derivation of Eq.\ \ref{eq:equivalence} follows from Proposition 1 
in \cite{Bach05}, and the fact that, given the ordered sequence of 
pivots $\mI$, the partial Cholesky factorization is unique. 

Using this factorization and the Woodbury identities (dropping the 
dependence on $\vc{\theta}$ and $\mI$ for clarity), the terms of the negative 
marginal log-likelihood (\ref{eq:pp_gp_nmll}) and variational free 
energy (\ref{eq:F}) become
\begin{align} 
E^{D} &\, =~ \sigma^{-2}\left(\vr{y}\vc{y} - \vr{y}
L\left(L^{\T}L + \sigtwo I\right)^{-1}L^{\T}\vc{y}
\right) \label{eq:L_data_fit} \\
E^C &\, = ~\log\left((\sigtwo)^{n-k}|L^{\T}L+
\sigtwo I|\right) \\
E^V &\, = ~\sigma^{-2}(\tr(K) - \tr(L^{\T}L)) \label{eq:E_v_exact}
\end{align}
%
%
%
%
We can further simplify the data term by augmenting the 
factor matrix as $\tL = [L^{\T},~ \sigma I_k]^{\T}$, 
where $I_k$ is the $k\!\times\! k$ identity matrix, and 
$\vc{\ty} = \trans{ [\trans{\vc{y}}, \trans{\vc{0}}_k]}$ is the $\vc{y}$ 
vector with $k$ zeroes appended:
\begin{equation}\label{eq:augmented_form} 
E^{D} = \sigma^{-2} 
\left( \vr{y}\vc{y} - \vr{\ty}\tL \, ( \tL^{\T}\tL )^{-1}\, \tL^{\T}\vc{\ty} \right) 
\end{equation}
Now, let $\tL = QR$ be a QR factorization of $\tL$, where $Q \in \Rm{(n+k)}{k}$ 
has orthogonal columns and $R \in \Rm{k}{k}$ is invertible. The first two terms in
the objective simplify further to
\begin{align} 
E^{D} 
&= \sigma^{-2}\left(\|\vc{y}\|^2 - \|Q^{\T}\vc{\ty}\|^2\right) \label{eq:augmented_data_fit_simplified} \\
E^C &=(n-k)\log(\sigtwo) + 2\log|R|\ .
\label{eq:augmented_complexity_simplified} 
\end{align}

\vspace*{-0.35cm} 
\subsection{Factorization update}
\label{subsec:factor_update}

\vspace*{-0.2cm}

Here we present the mechanics of the swap update algorithm,
see \cite{CAO2013} for pseudo-code.
Suppose we wish to swap inducing point $i$ with candidate point $j$ 
in $\mI_m$, the inducing set of size $m$. 
We first modify the factor matrices in order to remove
point $i$ from $\mI_m$, {\em i.e.} to downdate the factors.
Then we update all the key terms using one step of Cholesky and 
QR factorization with the new point $j$. 

Downdating to remove inducing point $i$ requires that we shift the 
corresponding columns/rows in the factorization to the right-most 
columns of $\tL$, $Q$, $R$ and to the last row of $R$.
We can then simply discard these last columns and rows, and modify 
related quantities. When permuting the order of the inducing points, the 
underlying GP model is invariant, but the matrices in the factored 
representation are not. If needed, any two points in $\mI_m$, 
can be permuted, and the Cholesky or QR factors can be updated in time
$O(mn)$. This is done with the efficient pivot permutation presented 
in the Appendix of \cite{Bach05}, with minor modifications to account 
for the augmented form of $\tL$.  In this way, downdating and removing 
$i$ take $O(mn)$ time, as does the updating with point $j$.

After downdating, we have factors $\tL_{m-1}$,$Q_{m-1}$, $R_{m-1}$, 
and inducing set $\mI_{m-1}$. To add $j$ to $\mI_{m-1}$, and update 
the factors to rank $m$,  one step of Cholesky factorization is 
performed with point $j$, for which, ideally, the new columnto append 
to $\tL$ is formed as
\begin{equation} 
\label{eq:exact_gk}
\! \bell_{m} = 
{(K\! -\!\hat{K}_{m-1})[:,j]} ~\Big/~ {\sqrt{(K\! -\!\hat{K}_{m-1})[j,j]}} 
\end{equation}
where $\hat{K}_{m-1} = L_{m-1} \trans{L_{m-1}}$. 
Then, we set $\tL_m = [\tL_{m-1} ~ \tilde{\bell}_m ]$, where
$\tilde{\bell}_m$ is just $\bell_m$ augmented
with $\sigma \be_m = [0,0,...,\sigma,...,0,0]^{\T}$.
The final updates are $Q_m = [ Q_{m-1} ~  \bq_m ]$, where $\bq_m$ is 
given by Gram-Schmidt orthogonalization,  
$\bq_m = ((I - Q_{m-1}Q_{m-1}^{\T})\tilde{\bell}_m) \, / \, {\|(I -
Q_{m-1}Q_{m-1}^{\T})\tilde{\bell}_m\|}$,
and $R_m$ is updated from $R_{m-1}$  so that $\tL_m = Q_mR_m$.

\vspace*{-0.15cm} 
\subsection{Evaluating candidates}
\vspace*{-0.2cm} 

Next we show how to select candidates for inclusion in the inducing 
set.  We first derive the exact change in the objective due to
adding an element to $\mI_{m-1}$.  Later we will provide an approximation
to this objective change that can be computed efficiently.

Given an inducing set $\mI_{m-1}$, and matrices 
$\tL_{m-1}, Q_{m-1}$, and $R_{m-1}$, we wish to evaluate the change 
in Eq.~\ref{eq:F} for $\mI_{m} \! =\!  \mI_{m-1} \cup j$.  That is,
$\Delta F \equiv F(\vc{\theta},\mI_{m-1})\!  -\!  F(\vc{\theta},\mI_{m})
= (\Delta E^D + \Delta E^C + \Delta E^V)/2$, where, based on the mechanics 
of the incremental updates above, one can show that
%
\begin{align} 
\Delta E^{D} &= \sigma^{-2}{(\vr{\ty}\left(I - Q_{m-1}Q_{m-1}^{\T}
  \right)\tilde{\bell}_m)^2} ~\Big/~ {\|\left(I - Q_{m-1}Q_{m-1}^{\T}
  \right)\tilde{\bell}_m\|^2} \label{exact_delta_d}\\
\Delta E^{C} &= \log\left(\sigtwo \right) -
\log\|(I - Q_{m-1}Q_{m-1}^{\T}
)\tilde{\bell}_m\|^2 \label{exact_delta_c}\\
\Delta E^{V} &= \sigma^{-2}\|\bell_m\|^2 \label{exact_delta_v}
\end{align}
This gives the exact decrease in the objective function after adding point $j$. 
For a single point this evaluation is $O(mn)$, so to evaluate all
$n-m$ points would be $O(mn^2)$.



\vspace*{-0.25cm} 
\subsubsection{Fast approximate cost reduction} 
\label{subsec:fast_cost}
\vspace*{-0.2cm} 

While $O(mn^2)$ is prohibitive, computing the exact change is not 
required.  Rather, we only need a ranking of the best few candidates.
Thus, instead of evaluating the change in the objective exactly, we 
use an efficient approximation based on a small number, $z$, of training
points which provide information about the residual between the 
current low-rank covariance matrix (based on inducing points) and 
the full covariance matrix. After this approximation proposes 
a candidate, we use the actual objective to decide whether to include it.
The techniques below reduce the complexity of evaluating all 
$n - m$ candidates to $O(zn)$.


To compute the change in objective for one candidate, we need the new 
column of the updated Cholesky factorization, $\bell_m$.
In Eq.~(\ref{eq:exact_gk}) this vector is a (normalized) column of the 
residual $K\!-\! \hat{K}_{m-1}$ between the full kernel matrix and the
Nystr\"{o}m approximation.  Now consider the full Cholesky decomposition 
of $K = L^{*} L^{*\T}$ where $L^{*} = [L_{m-1}, L(\mJ_{m-1})]$ is 
constructed with $\mI_{m-1}$ as the first pivots and 
$\mJ_{m-1} = \{1,...,n\} \backslash \mI_{m-1}$ as the remaining pivots, 
so the residual becomes $K\!-\! \hat{K}_{m-1} = L(\mJ_{m-1})L(\mJ_{m-1})^\T$.
We approximate $L(\mJ_{m-1})$ by a rank $z \!\ll\! n$ matrix, 
$L_z$, by taking $z$ points from $\mJ_{m-1}$ and performing a partial 
Cholesky factorization of $K\! -\! \hat{K}_{m-1}$ using these pivots.
The residual approximation becomes $K\!-\!\hat{K}_{m-1} \approx L_z L_z^\T$,
and thus $\bell_{m} \approx (L_{z}L_{z}^{\T})[:,j] \Big/ \sqrt{(L_{z}L_{z}^{\T})[j,j]}$.
The pivots used to construct $L_z$ are called {\em information pivots}; 
their selection is discussed in Sec.~\ref{subsec:info_pivot_selection}.

The approximations to $\Delta E^D_k$, $\Delta E^C_k$ and $\Delta E^V_k$,
Eqs.~(\ref{exact_delta_d})-(\ref{exact_delta_v}), for all candidate points, 
involve the following terms:
$\diag(L_zL_z^{\T}L_zL_z^{\T})$, $\vr{y}L_zL_z^{\T}$, and $\left( Q_{k-1}[1:n,:]\right)^{\T} L_zL_z^{\T}$. 
The first term can be computed in time $O(z^2n)$, and the other two in 
$O(z m n)$ with careful ordering of matrix multiplications.\footnote{Both 
can be further reduced to $O(z n)$ by appropriate caching during 
the updates of $Q$,$R$ and $\tL$, and $L_z$} 
Computing $L_z$ costs $O(z^2n)$, but can be avoided since information 
pivots change by at most one when an information pivots is added to 
the inducing set and needs to be replaced.  The techniques in 
Sec.~\ref{subsec:factor_update} bring the associated update cost 
to $O(zn)$ by updating $L_z$ rather than recomputing it.  
These $z$ information pivots are equivalent to the ``look-ahead'' 
steps of Bach and Jordan's CSI algorithm, but as described in 
Sec.~\ref{subsec:info_pivot_selection}, there is a more effective way 
to select them.

\vspace*{-0.25cm} 
\subsubsection{Ensuring a good approximation} 
\label{subsec:info_pivot_selection}
\vspace*{-0.2cm} 

\begin{wrapfigure}{r}{0.5\textwidth}
\vspace*{-0.6cm}
\subfigure{
\includegraphics[width=0.25\textwidth]{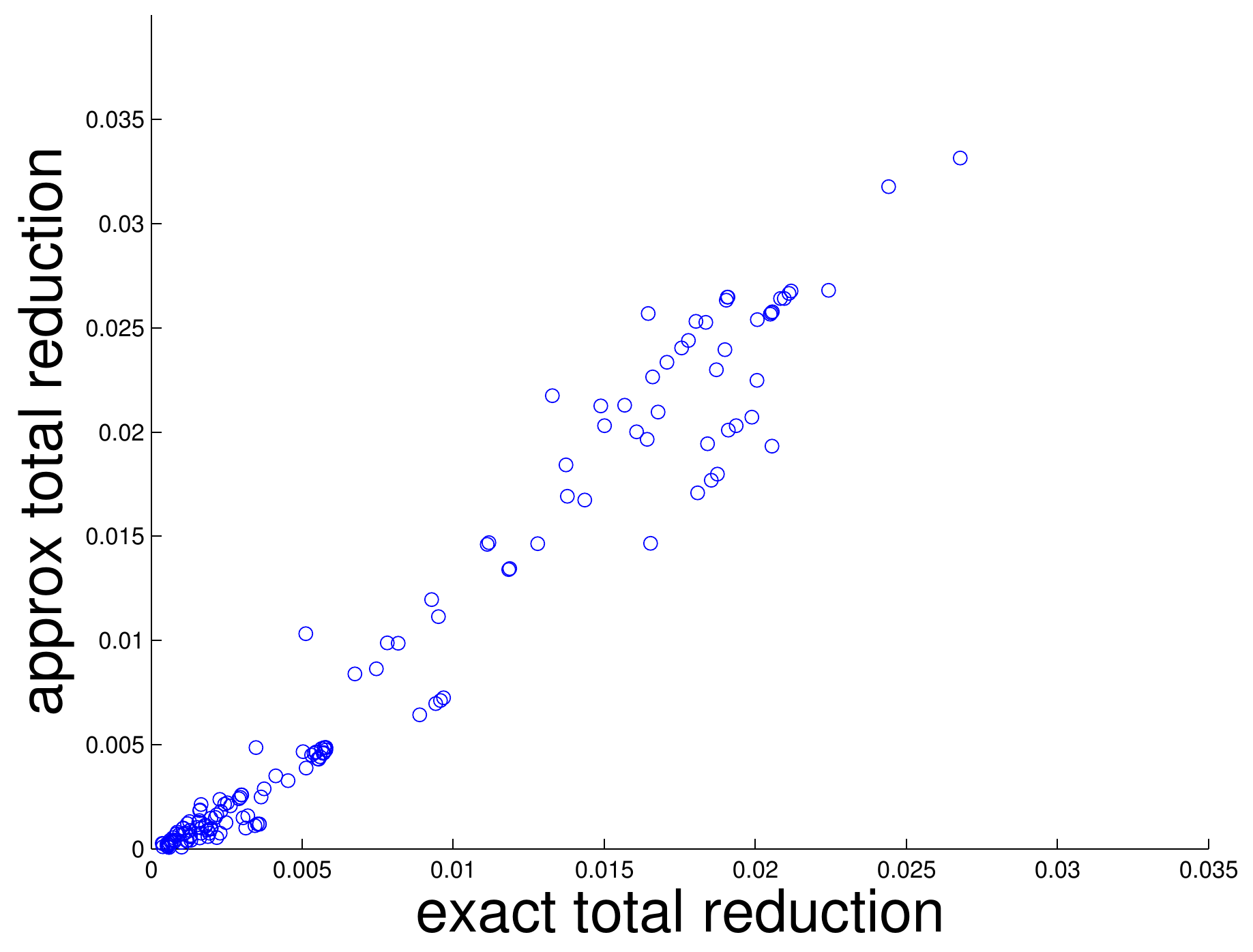} 
}%
\subfigure{
\includegraphics[width=0.25\textwidth]{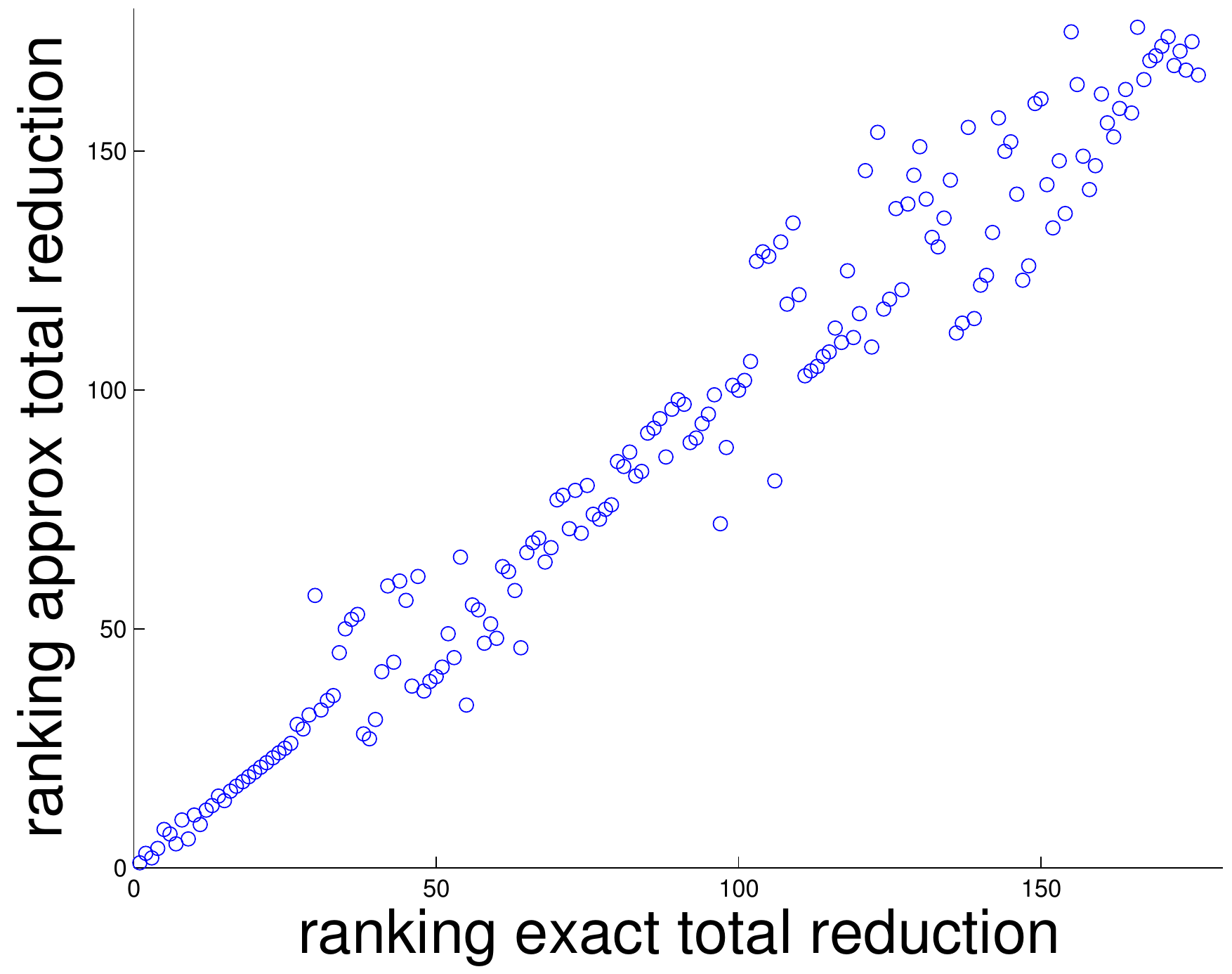}
}%

\vspace*{-0.4cm}
\caption{Exact vs approximate costs, based on the 1D example of
Sec.\ \ref{sec:experiments}, with $z\! = \!10$, $n\!=\!200$.
\label{fig:costs}
}%
\vspace*{-0.4cm}
\end{wrapfigure}

\comment{{{
New story in this part: 1. randomly selecting info pivots
work just as well on average, as long as reshuffling is done in the
  same way; 2. can get significant initial speed-up by starting with a
very small info-pivot set, and increase it until $z$ only when 
approximation is bad, however, this generally hurts model quality a bit


We greedily choose the information pivots based on the trace
norm of the approximation residual.
Let $\mI_{z-1}$ and $\mI_{z}$ denote the index set
before and after inclusion of one new such pivot.
This score is the difference between  $\tr(K - \hat{K}_{k-1} - L(\mI_{z-1})L(\mI_{z-1})^\T)$,
and $\tr(K - \hat{K}_{k-1} - L(\mI_{z})L(\mI_{z})^\T)$, which is
$\tr(L(\mI_{z})L(\mI_{z})^\T - L(\mI_{z-1})L(\mI_{z-1})^\T) = \|\bell_{z}\|^2$,
where $\bell_{z}$ is the new column obtained via one step of partial Cholesky
using the new information pivot $i_{z}$.
However, directly computing this would be expensive as it would
require one step of Cholesky factorization, instead
we use the lower bound, $\bell_{z}[i_{z}]^2 \le
\|\bell_{z}\|^2$, which according to Eq.\,(\ref{alg:cholesky_step1}), is
$\vc{d}^{z}[i_{z}]$.
In the context of covariance matrices these $\vc{d}$ values, which 
are maintained during partial Cholesky steps, are also the variance at 
each point that has not been explained by the low rank decomposition.
That is, we successively select the points with largest ``unexplained"
point-wise variance to be information pivots. 
This criteria is closely related to the maximum differential
entropy criteria used in the IVM \cite{Lawrence03}.

As the set of inducing points is updated the approximation provided
by the set of information pivots may naturally degrade. Although an
information pivot could itself be added to the inducing set, forcing
slight update to the information pivot set, in practice, it is
beneficial to occasionally update the whole information pivot set
($O(z^2n)$ operation) to ensure good approximation. In practice, we
find that a small predetermined number of times is sufficient, while
the when and how it is decided to be done has little impact on the
test performance. 
Overall, our algorithm is fairly robust to the quality of this
approximation by information pivots, because all we care about is the
relative ranking of a few best candidates. This is illustrated in
Fig.\,\ref{fig:costs}.
}}}

Selection of the information pivots determines the approximate objective, 
and hence the candidate proposal. To ensure a good approximation, 
the CSI algorithm \cite{Bach05} greedily selects points to find an 
approximation of the residual $K\! -\! \hat{K}_{m-1}$ in 
Eq.\,(\ref{eq:exact_gk}) that is optimal in terms of a bound of the trace norm.
The goal, however, is to approximate 
Eqs.\,(\ref{exact_delta_d})-(\ref{exact_delta_v})\,. 
By analyzing the role of the residual matrix, we see that the information 
pivots provide a low-rank approximation to the orthogonal complement of
the space spanned by current inducing set. 
With a fixed set of information pivots, parts of that subspace may never 
be captured.  This suggests that we might occasionally update the entire 
set of information pivots.
Although information pivots are changed when one is moved into the inducing 
set, we find empirically that this is not insufficient.  Instead, at regular 
intervals we replace the entire set of information pivots by random 
selection.  We find this works better than optimizing the information
pivots as in \cite{Bach05}.  

Figure \ref{fig:costs} compares the exact and approximate cost reduction for candidate inducing 
points (left), and their respective rankings (right). The
approximation is shown to work well.  It is also robust to changes 
in the number of information pivots and the frequency of updates.
When bad candidates are proposed, they are rejected 
after evaluating the change in the true objective.
We find that rejection rates are typically low during early iterations 
($< 20\%$), but increase as optimization nears convergence 
(to $30\%$ or $40\%$).
Rejection rates also increase for sparser models, where each inducing 
point plays a more critical role and is harder to replace.

\vspace*{-0.2cm} 
\subsection{Hybrid optimization}
\label{sec:Hyperparams}
\vspace*{-0.25cm} 

The overall hybrid optimization procedure performs block 
coordinate descent in the inducing points and the continuous hyperparameters. 
It alternates between discrete and continuous phases until improvement in 
the objective is below a threshold or the computational time budget is 
exhausted. 

In the discrete phase, inducing points are considered for swapping 
with the hyper-parameters fixed.  With the factorization and efficient 
candidate evaluation above, swapping an inducing point $i \in \mI_m$ 
proceeds as follows: 
\begin{inparaenum}[(I)]
\item down-date the factorization matrices as in
Sec.~\ref{subsec:factor_update} to remove $i$;
\item compute the true objective function value $F_{m-1}$ over the down-dated 
model with $\mI_m \backslash \{i\}$, using (\ref{eq:augmented_data_fit_simplified}), (\ref{eq:augmented_complexity_simplified}) and (\ref{eq:E_v_exact});
\item select a replacement candidate using the fast approximate cost
change from Sec.~\ref{subsec:fast_cost}; 
\item evaluate the exact objective change, using (\ref{exact_delta_d}), 
(\ref{exact_delta_c}), and (\ref{exact_delta_v}); 
\item add the exact change to the true objective $F_{m-1}$ to get 
the objective value with the new candidate.  
If this improves, we include the candidate in $\mI$ and update the 
matrices as in Sec.~\ref{subsec:factor_update}.  Otherwise it is rejected
and we revert to the factorization with $i$; 
\item if needed, update
the information pivots as in Secs.~\ref{subsec:fast_cost} and
\ref{subsec:info_pivot_selection}. 
\end{inparaenum}

After each discrete optimization step we fix the inducing set $\mI$ 
and optimize the hyperparameters using non-linear conjugate gradients (CG).
The equivalence in (\ref{eq:equivalence}) allows us to compute the 
gradient with respect to the hyperparameters analytically using 
the Nystr\"{o}m form. 
In practice, because we alternate each phase for many training
epochs, attempting to swap every inducing point in each epoch is 
unnecessary, just as there is no need to run hyperparameter optimization 
until convergence.
As long as all inducing set points are eventually considered
we find that optimized models can achieve similar performance
with shorter learning times.

\vspace*{-0.15cm} 
\section{Experiments and analysis}
\label{sec:experiments}
\vspace*{-0.2cm} 

For the experiments that follow we jointly learn inducing points 
and hyperparameters, a more challenging task than learning inducing 
points with known hyperparameters \cite{Seeger03,Snelson06}. For 
all but the 1D example, the number of inducing points swapped per epoch
is $min(60,m)$.  The maximum number of function evaluations per epoch in 
CG hyperparameter optimization is $min(20, max(15,2d))$, where $d$ is 
the number of continuous hyperparameters.  Empirically we find the algorithm
is robust to changes in these limits. We use two performance 
measures, (a) standardized mean square error (SMSE), 
$\frac{1}{N}\Sigma_{t=1}^{N}(\hat{y}_{t} - y_t)^2/\hat{\sigma}^{2}_{*}$, 
where $\hat{\sigma}^{2}_{*}$ is the sample variance of test outputs $\{y_t\}$, 
and (2) standardized negative log probability (SNLP) defined 
in \cite{rasmussen2006gaussian}. 
\vspace*{-0.2cm} 

\subsection{Discrete input domain}
\label{subsec:discrete_domain_exp}

\vspace*{-0.25cm} 
\begin{figure*}[t]%

\begin{center}

\vspace*{-0.3cm}

\subfigure{
\includegraphics[width=0.33 \textwidth]{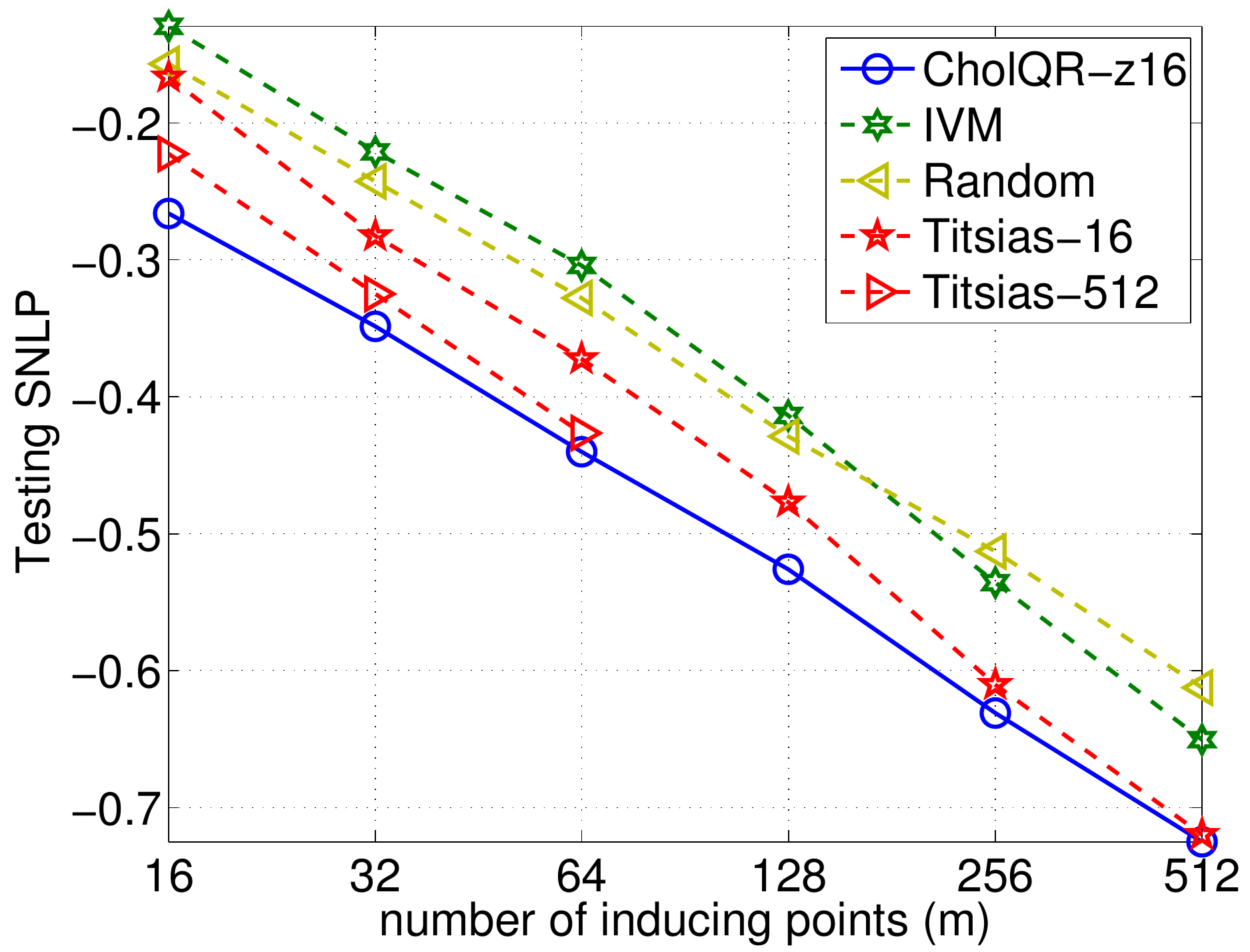}
\label{fig:graph_curves_snlp}
}%
\subfigure{
\includegraphics[width=0.33\textwidth]{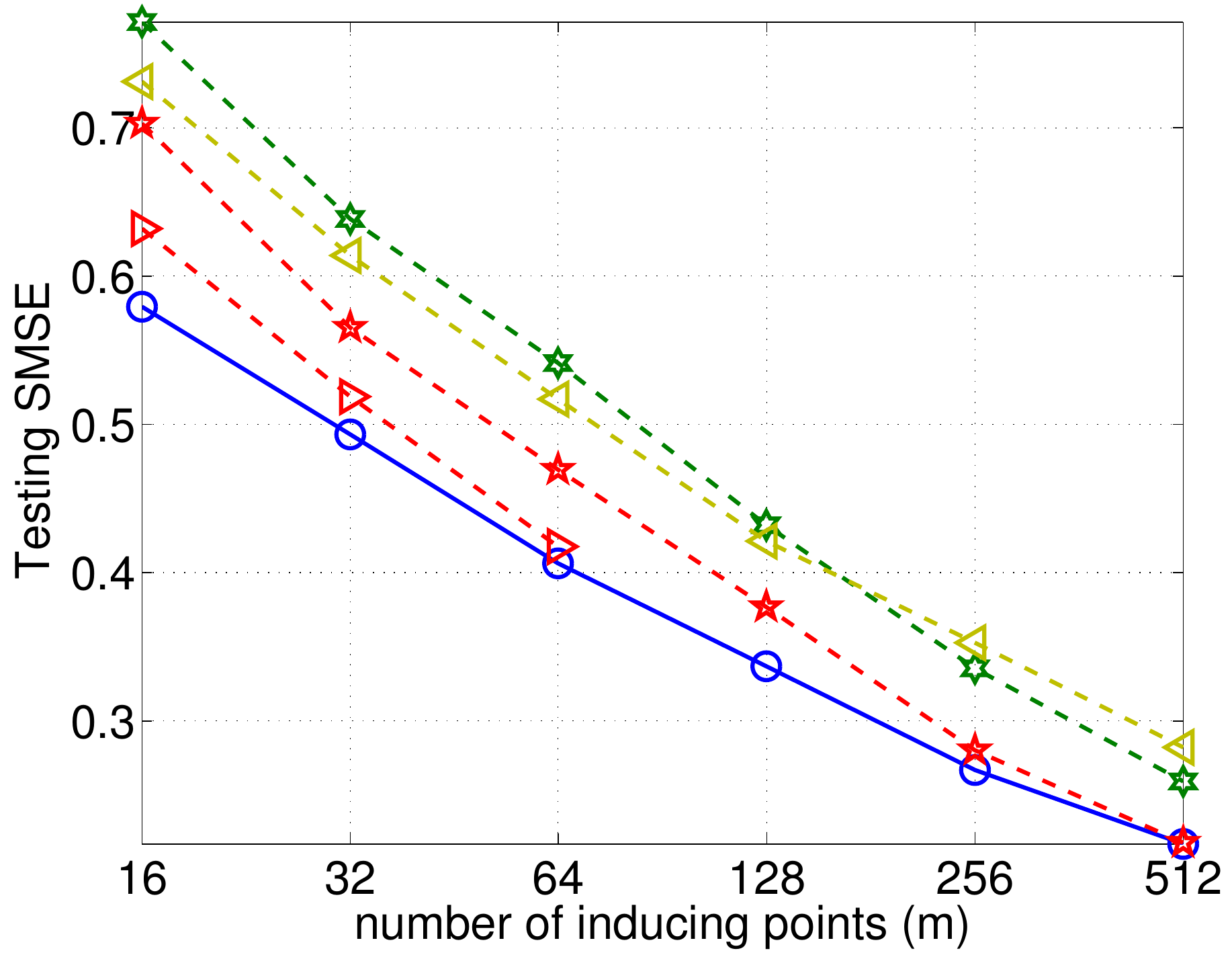}
\label{fig:graph_curves_smse}
}%

\vspace*{-0.25cm}

\subfigure{
\includegraphics[width=0.33\textwidth]{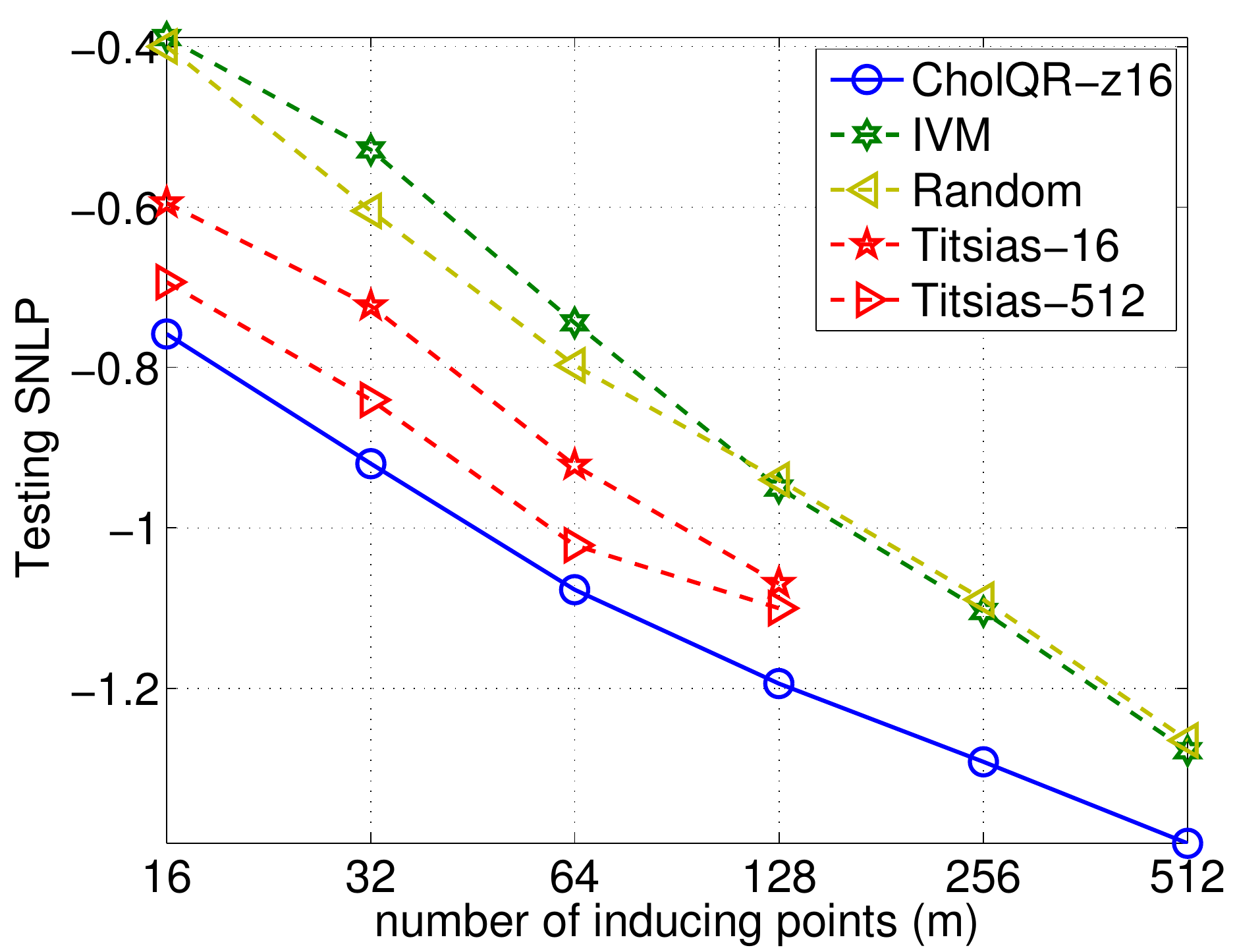}
\label{fig:graph_curves_snlp}
}%
\subfigure{
\includegraphics[width=0.33\textwidth]{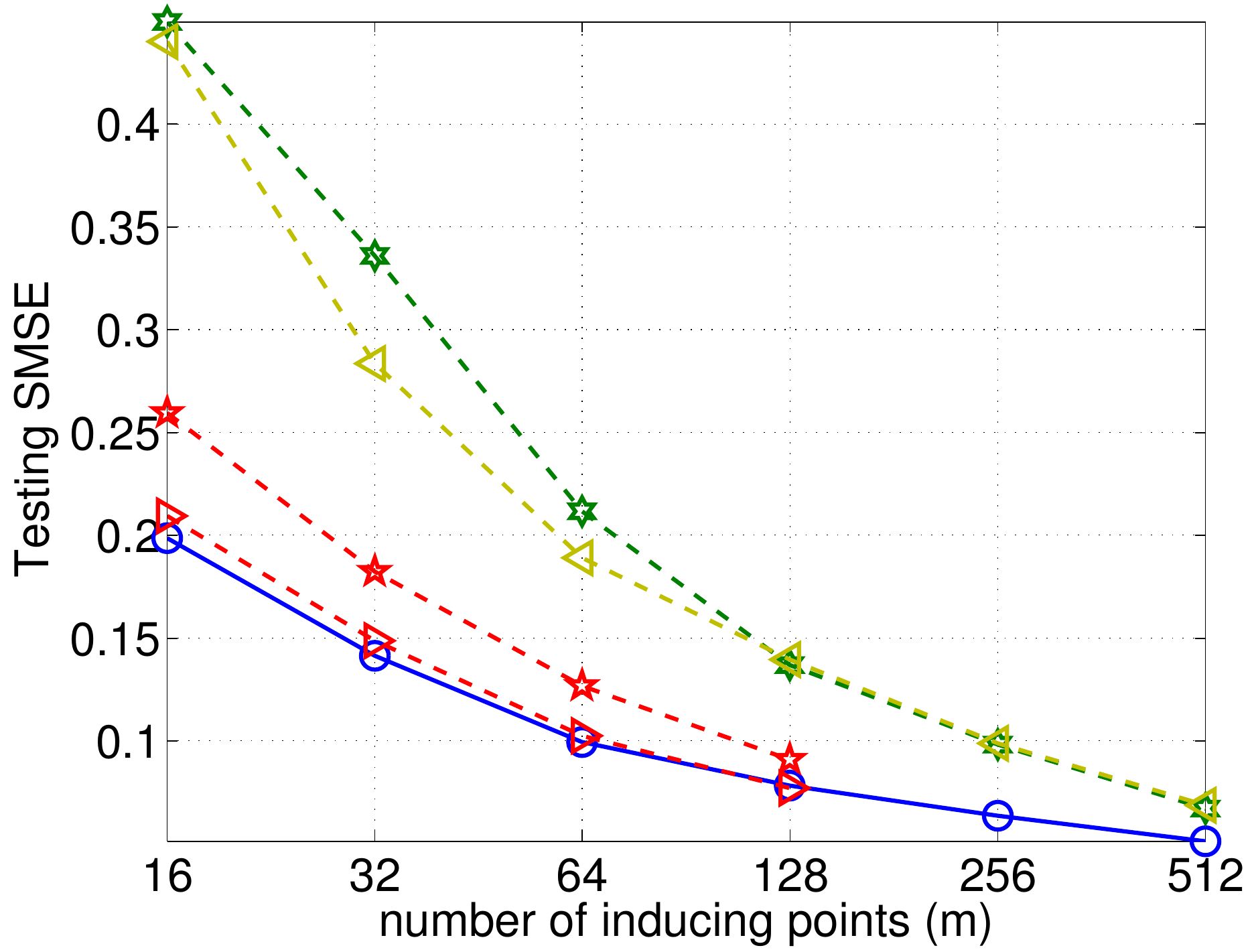}
\label{fig:graph_curves_smse}
}%
\end{center}

\vspace*{-0.15cm}
\caption{Test performance on discrete datasets.
\textbf{(top row)} BindingDB, values at each marker is the average of 150 runs (50-fold random train/test splits times 3 random initialization);
\textbf{(bottom row)} HoG dataset, each marker is the average of 10 randomly initialized runs. 
} 
\label{fig:bindingDB_curves}

\vspace*{-0.3cm}
\end{figure*}

We first show results on two discrete datasets with kernels 
that are not differentiable in the input variable $x$.
Because continuous relaxation methods are not applicable, we compare
to discrete selection methods, namely, random selection as baseline 
(Random), greedy subset-optimal selection of Titsias \cite{Titsias09} 
with either 16 or 512 candidates (Titsias-16 and Titsias-512),
and Informative Vector Machine \cite{Lawrence03} (IVM).
For learning continuous hyperparameters, each method optimizes 
the same objective using non-linear CG.
Care is taken to ensure consist initialization and termination
criteria \cite{CAO2013}.
For our algorithm we use $z=16$ information pivots with random selection
(CholQR-z16).  Later, we show how variants of our algorithm trade-off speed
and performance.
{\comment{Further, since it is not our goal to study which one of PP likelihood 
or variational free energy is the better objective function, we use the 
variational objective for all methods, and hint at some of the differences 
in the next section.}
Additionally, we also compare to least-square kernel regression using CSI
(in Fig.\,\ref{fig:trade_off_all}).


The first discrete dataset, from \url{bindingdb.org}, concerns the 
prediction of binding affinity for a target (Thrombin), from the 2D 
chemical structure of small molecules (represented as graphs).
We do 50-fold random splits to 3660 training points and 192 test 
points for repeated runs.  We use a compound kernel, comprising
14 different 
graph kernels, and 15 continuous hyperparameters (one noise variance 
and 14 data variances).
In the second task, from \cite{BoS10}, the task is to predict 3D human 
joint position from histograms of HoG image features \cite{DalalTriggs05}.
Training and test sets have $4819$ and $4811$ data points.
Because our goal is the general purpose sparsification method for GP
regression, we make no attempt at the more difficult problem
of modelling the multivariate output structure in the regression 
as in \cite{BoS10}.
Instead, we predict the vertical position of joints independently,
using a histogram intersection kernel \cite{libpmk}, having
four hyperparameters: one noise variance, and three data 
variances corresponding to the kernel evaluated over the HoG from each 
of three cameras. 
We select and show result on the representative left wrist here
(see \cite{CAO2013} for others joints, and more details about the
datasets and kernels used).

The results in Fig.\ \ref{fig:bindingDB_curves} and \ref{fig:cost_time}
show that CholQR-z16 outperforms the baseline methods in terms of 
test-time predictive power with significantly lower training time.
Titsias-16 and Titsias-512 shows similar test performance, but they
are two to four orders of magnitude slower than CholQR-z16
(see Figs.\ \ref{fig:graph_time_snlp} and \ref{fig:graph_time_smse}).
Indeed, Fig.\ \ref{fig:graph_curves_total_time} shows that the training
time for CholQR-z16 is comparable to IVM and Random selection, but with 
much better performance.
The poor performance of Random selection highlights the importance of 
selecting good inducing points, as no amount of hyperparameter optimization 
can correct for poor inducing points. 
Fig.\ \ref{fig:graph_curves_total_time} also shows IVM
to be somewhat slower due to the increased number of iterations needed, 
even though per epoch, IVM is faster than CholQR.  When stopped earlier, IVM
test performance further degrades.

\begin{figure*}[t]%

\vspace*{-0.3cm}

\subfigure[][]{
\includegraphics[width=0.32\textwidth]{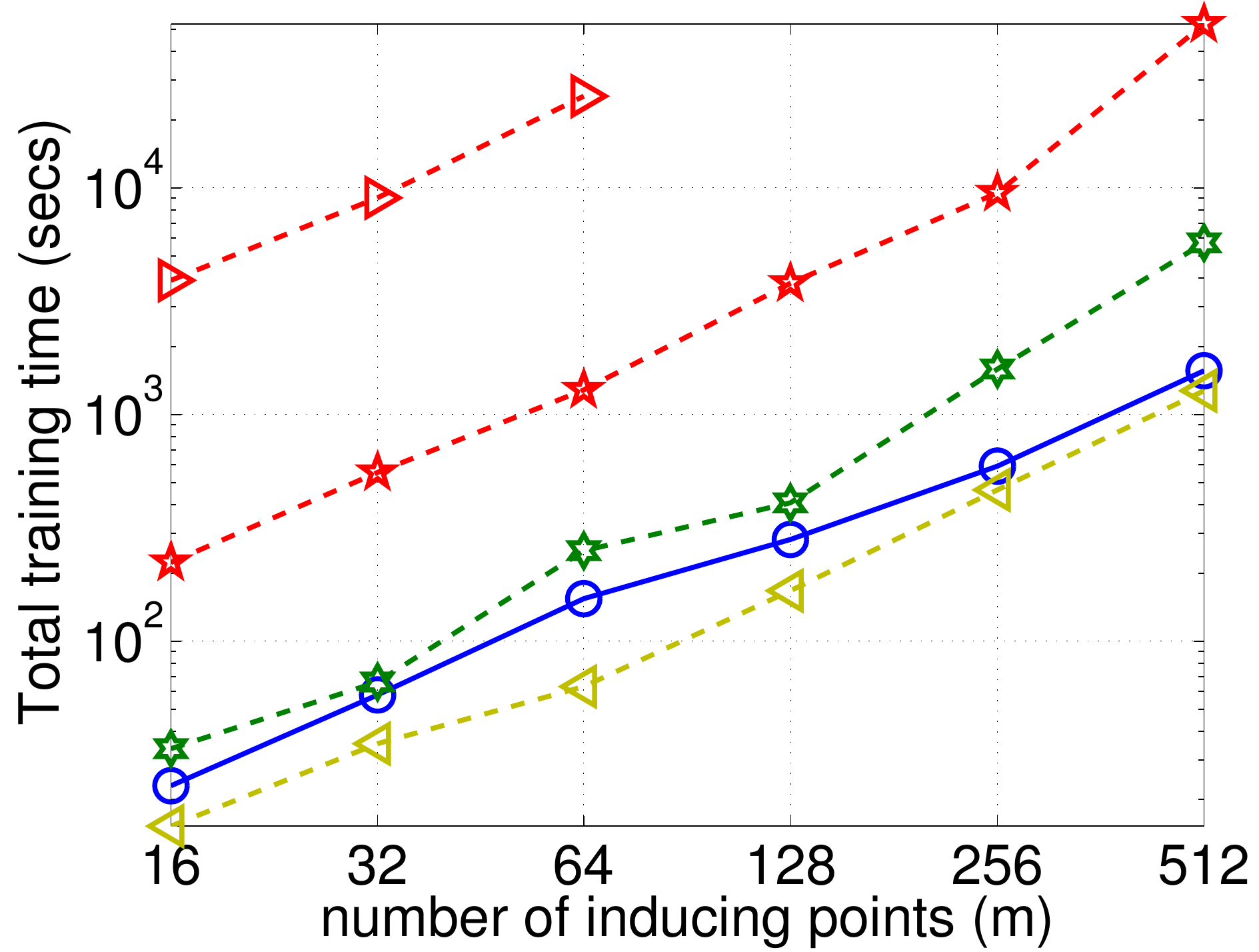}
\label{fig:graph_curves_total_time}
}%
\subfigure[][]{
\includegraphics[width=0.32\textwidth]{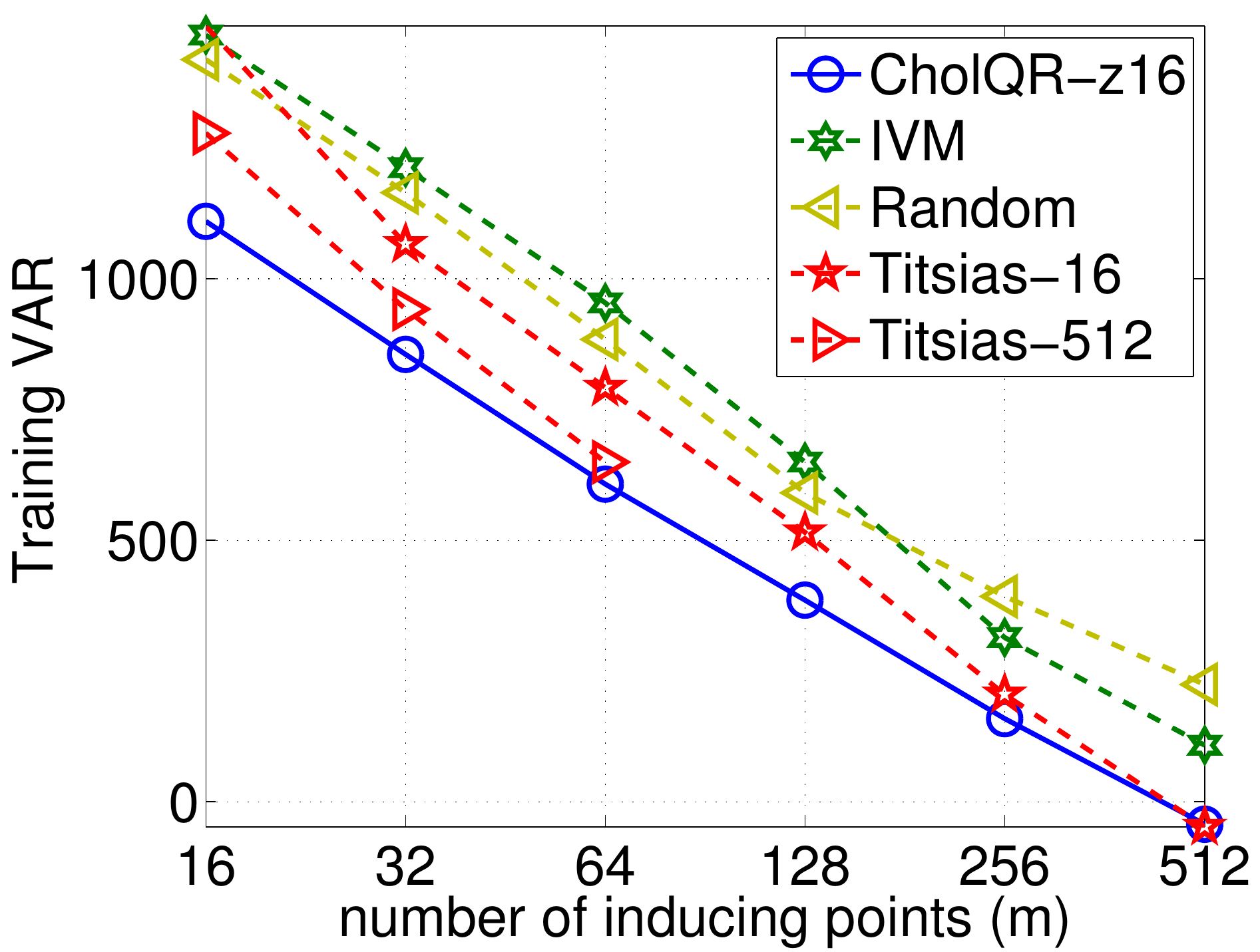}
\label{fig:graph_curves_var}
}%
\subfigure[][]{
\includegraphics[width=0.32\textwidth]{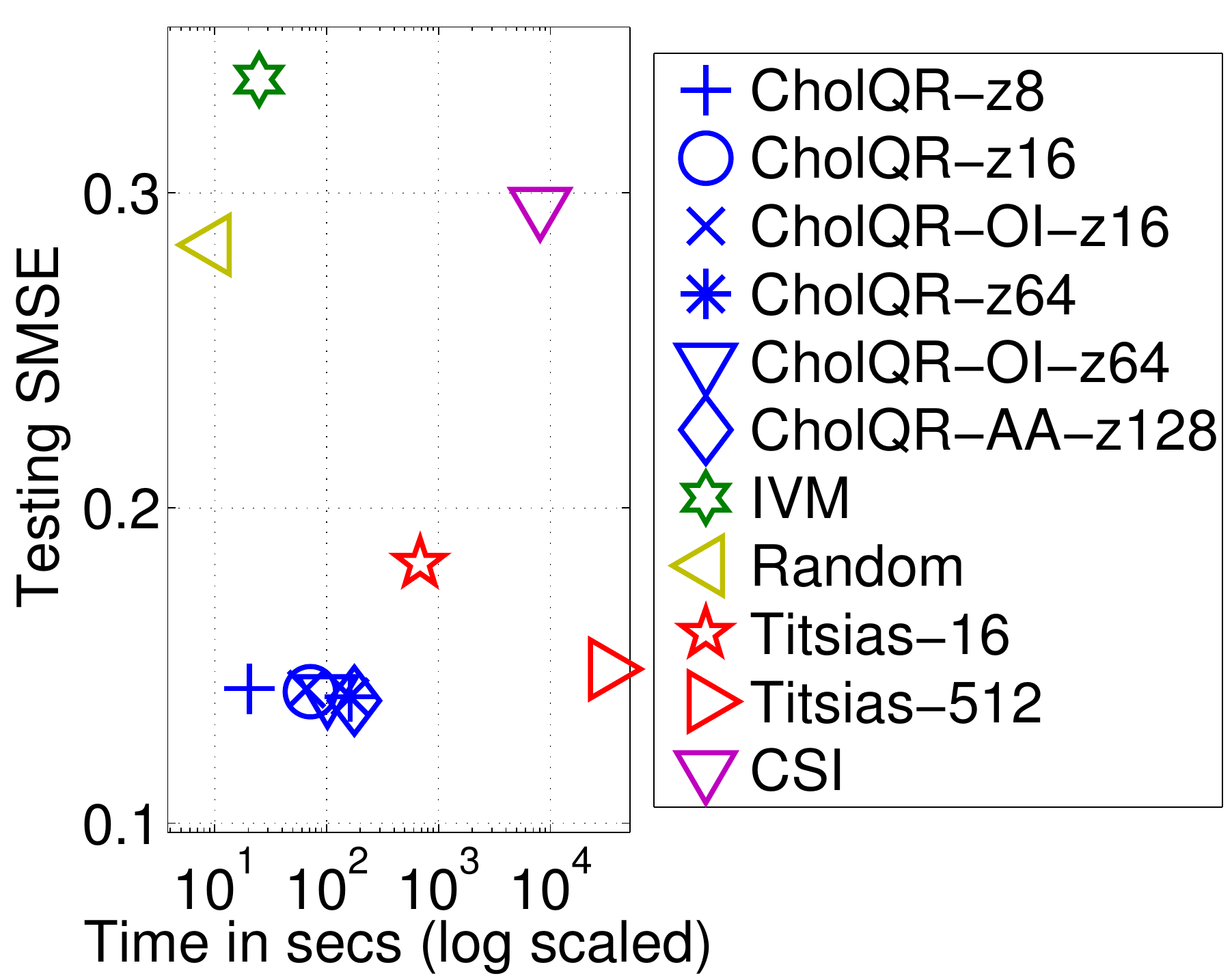}
\label{fig:trade_off_all}
}%

\vspace*{-0.35cm}
\begin{center}
\subfigure[][]{
\includegraphics[width=0.32\textwidth]{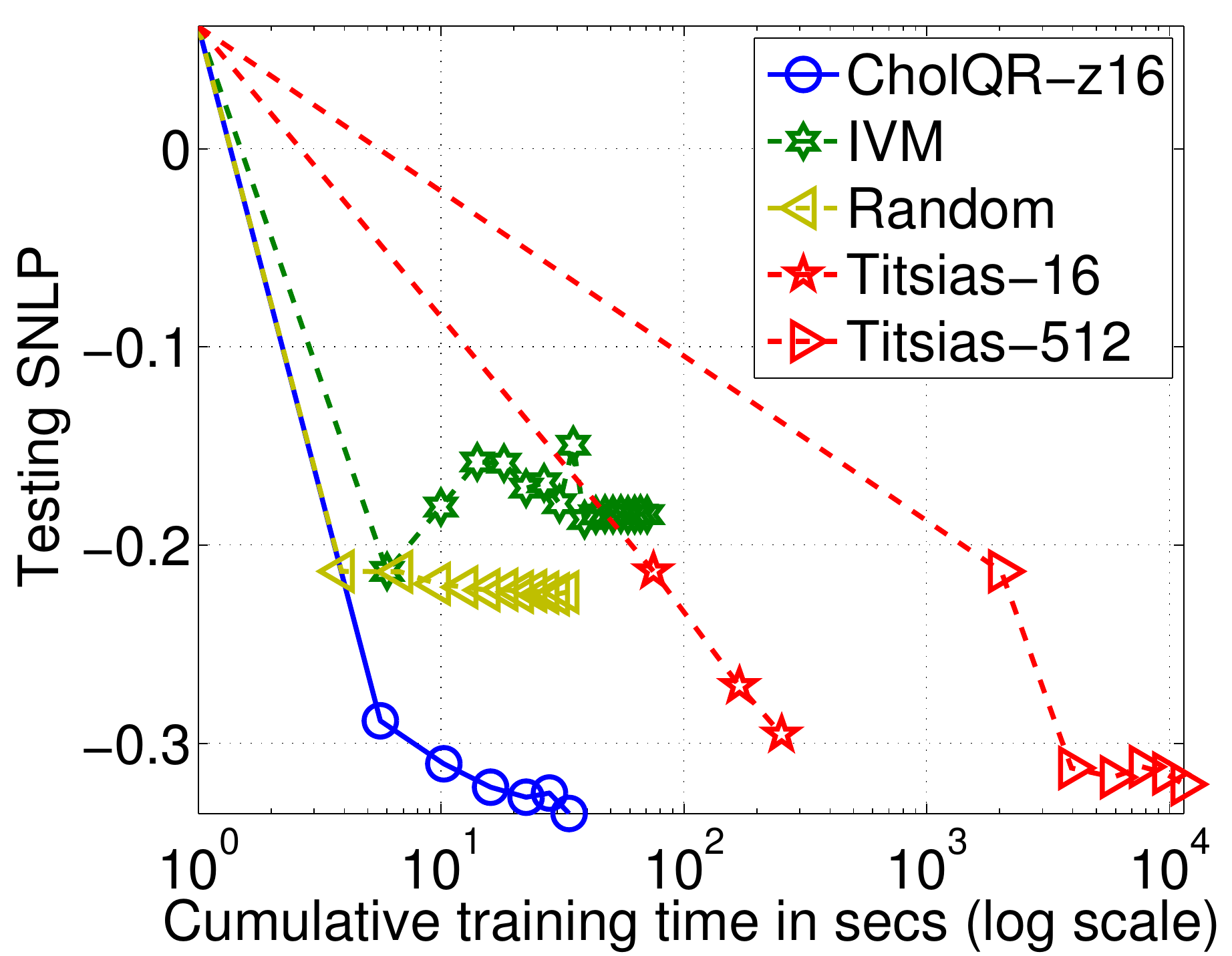}
\label{fig:graph_time_snlp}
}%
\subfigure[][]{
\includegraphics[width=0.32\textwidth]{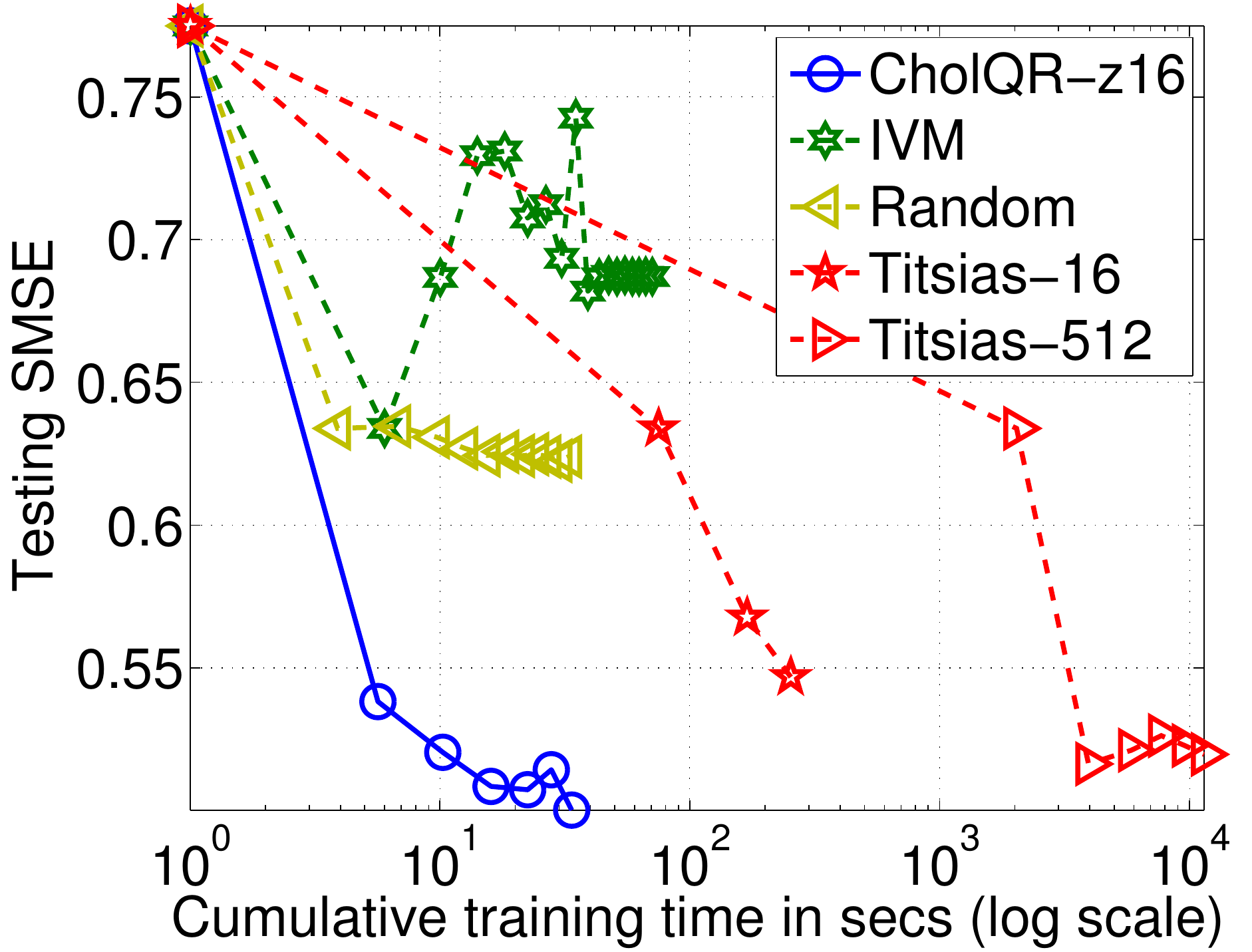}
\label{fig:graph_time_smse}
}%
\subfigure[][]{
\includegraphics[width=0.32\textwidth]{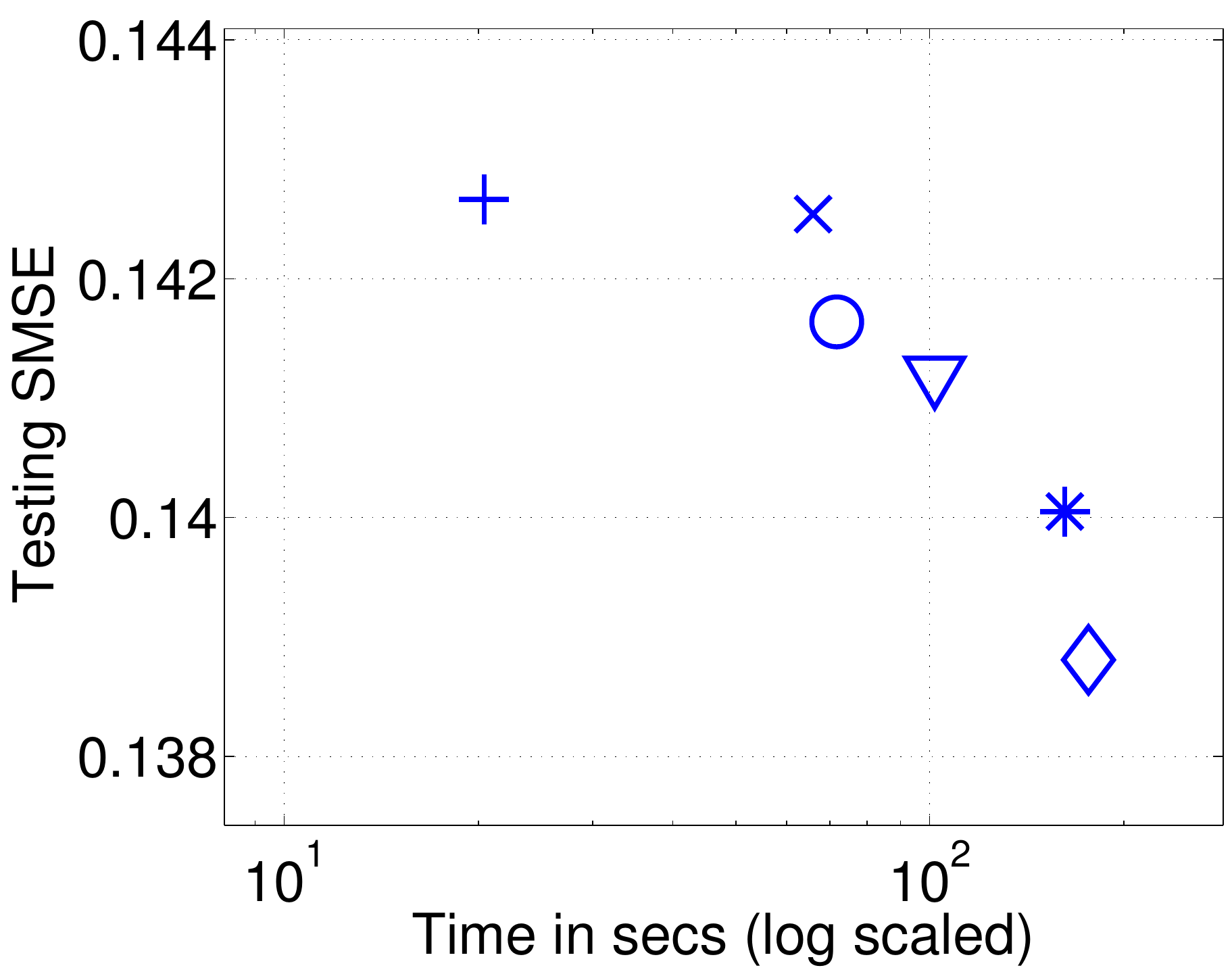}
\label{fig:trade_off_cholqr}
}%
\end{center}

\vspace*{-0.3cm}
\caption{Training time versus test performance on discrete datasets. 
\textbf{\subref{fig:graph_curves_total_time}} the average BindingDB 
training time; \textbf{\subref{fig:graph_curves_var}} the average BindingDB
objective function value at convergence;
\textbf{\subref{fig:graph_time_snlp}} and \textbf{\subref{fig:graph_time_smse}}
show test scores versus training time with $m=32$ for a single run;
\textbf{\subref{fig:trade_off_all}} shows the trade-off between training
time and testing SMSE on the HoG dataset with $m=32$, for various methods
including multiple variants of CholQR and CSI;
\textbf{\subref{fig:trade_off_cholqr}} a zoomed-in version of
\subref{fig:trade_off_all} comparing the variants of CholQR.
  }
\label{fig:cost_time}

\vspace*{-0.3cm}

\end{figure*}

Finally, Fig.\ \ref{fig:trade_off_all} and \ref{fig:trade_off_cholqr} 
show the trade-off between the test SMSE and training time for variants 
of CholQR, with baselines and CSI kernel regression \cite{Bach05}.
For CholQR we consider different numbers of information pivots (denoted
z8, z16, z64 and z128), and different strategies for their selection
including random selection, optimization as in \citep{Bach05} (denote OI)
and adaptively growing the information pivot set (denoted AA, see
\cite{CAO2013} for details).
These variants of CholQR trade-off speed and performance
(\ref{fig:trade_off_cholqr}), all significantly outperform
the other methods (\ref{fig:trade_off_all}); CSI, which uses grid 
search to select hyper-parameters, is slow and exhibits higher SMSE.

\vspace*{-0.2cm} 
\subsection{Continuous input domain}
\vspace*{-0.25cm} 

\begin{figure*}[t]%

\centering

\vspace*{-0.2cm}
\subfigure[][CholQR-MLE]{
\includegraphics[width=0.18\textwidth]{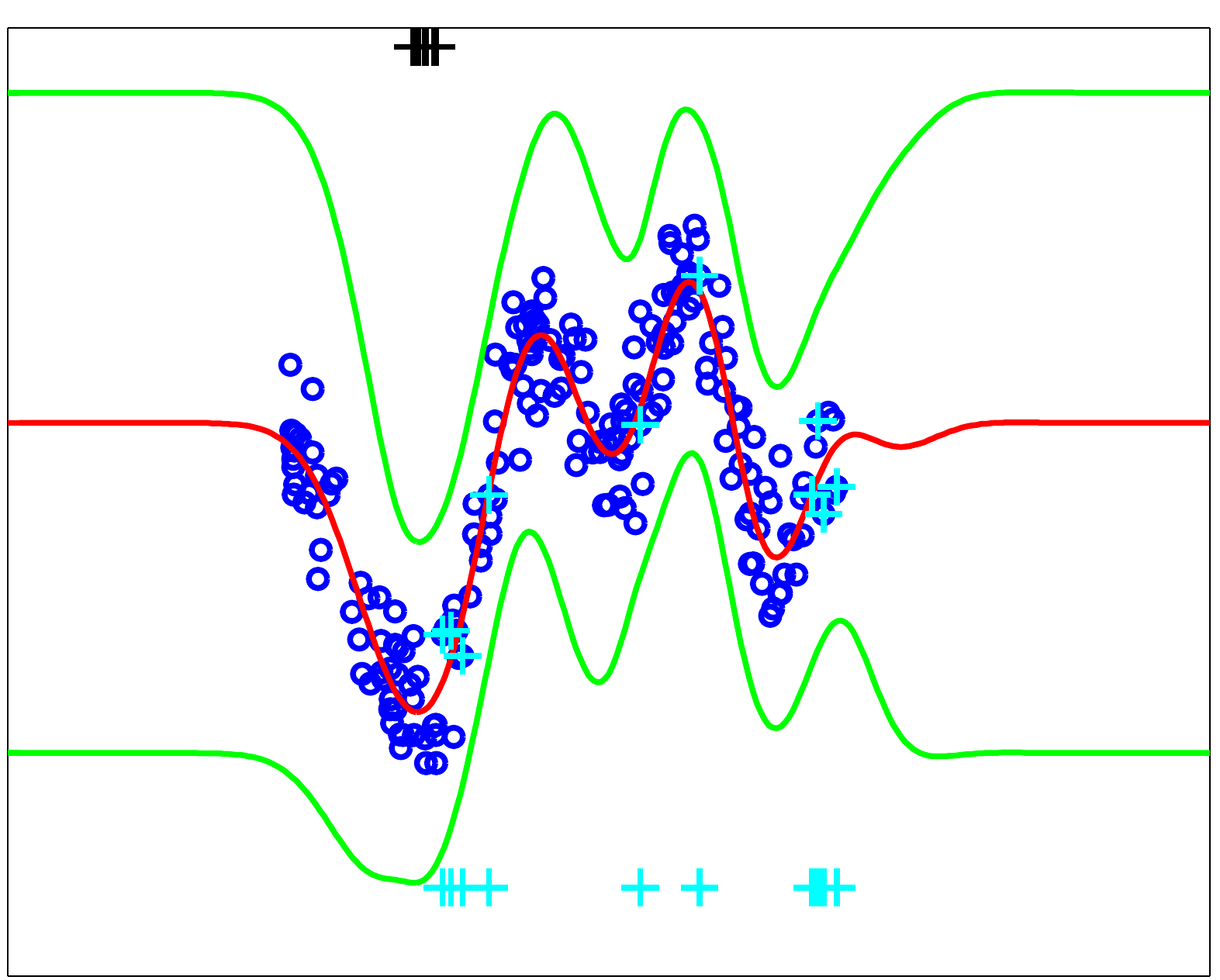}
\label{fig:cholqr_mle_200}
}%
\subfigure[][CholQR-MLE]{
\includegraphics[width=0.18\textwidth]{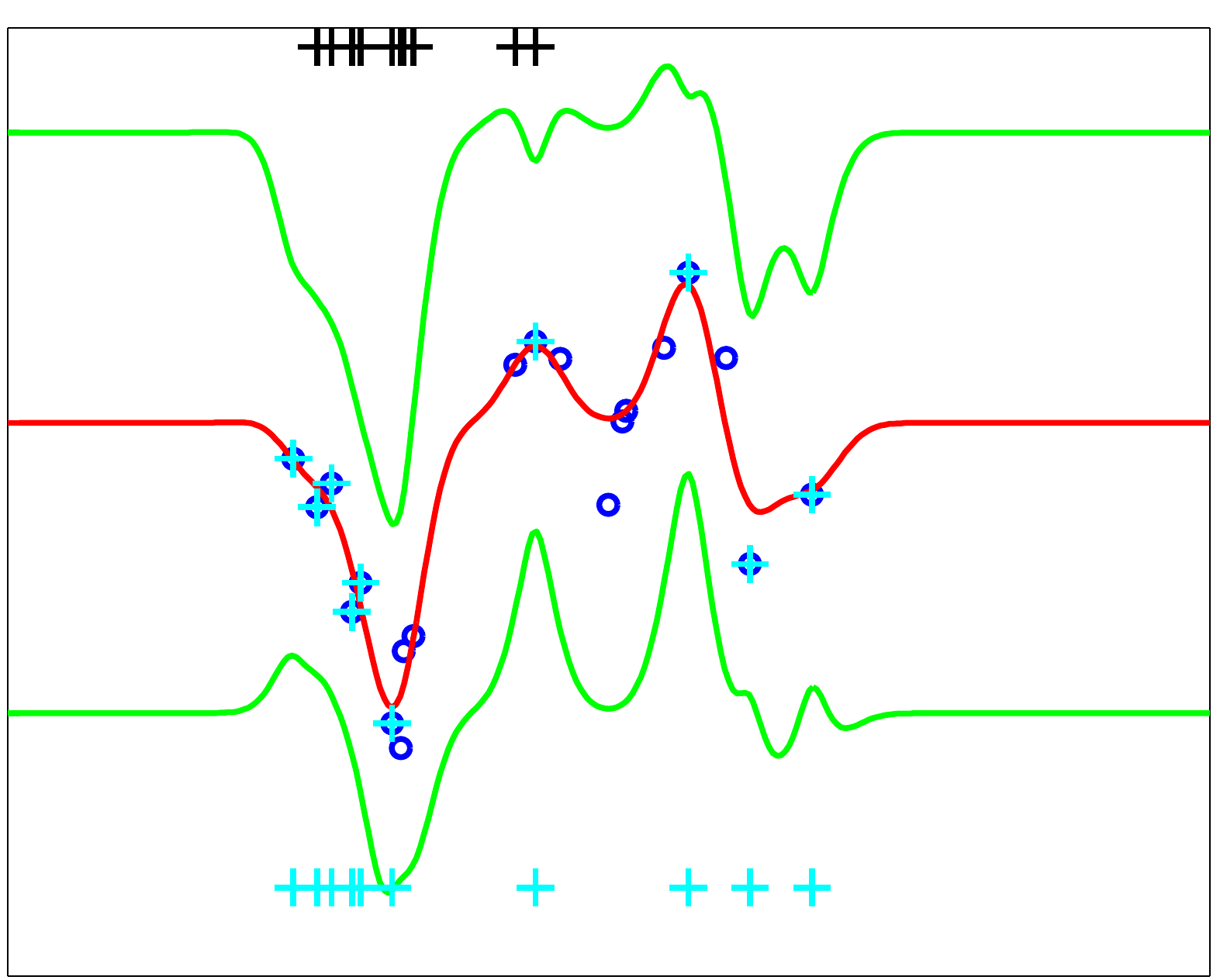}
\label{fig:cholqr_mle_20}
}%
\subfigure[][SPGP]{
\includegraphics[width=0.18\textwidth]{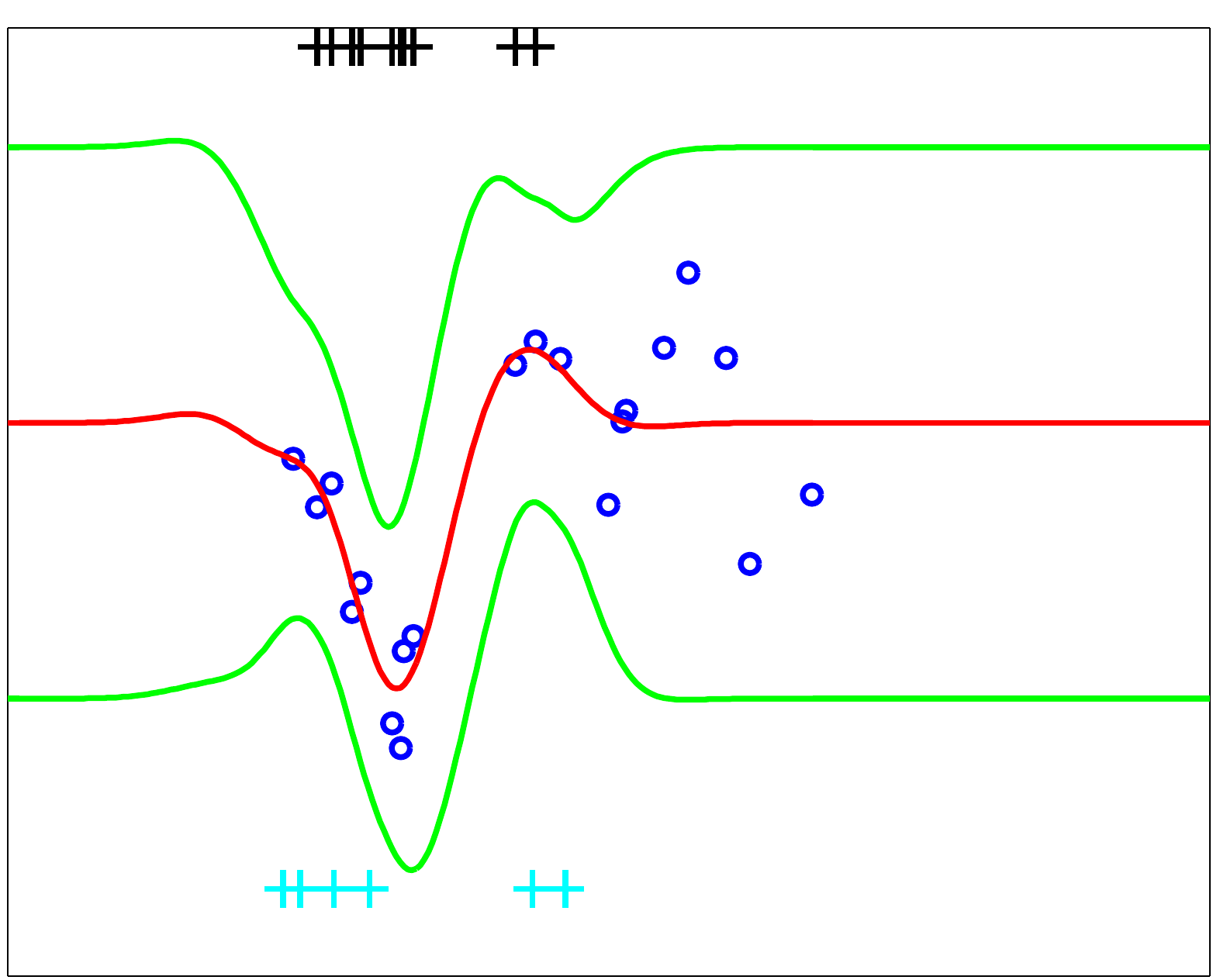}
\label{fig:fitc_20_clumped_ini}
}%

\vspace*{-0.195cm}
\subfigure[][CholQR-VAR]{
\includegraphics[width=0.18\textwidth]{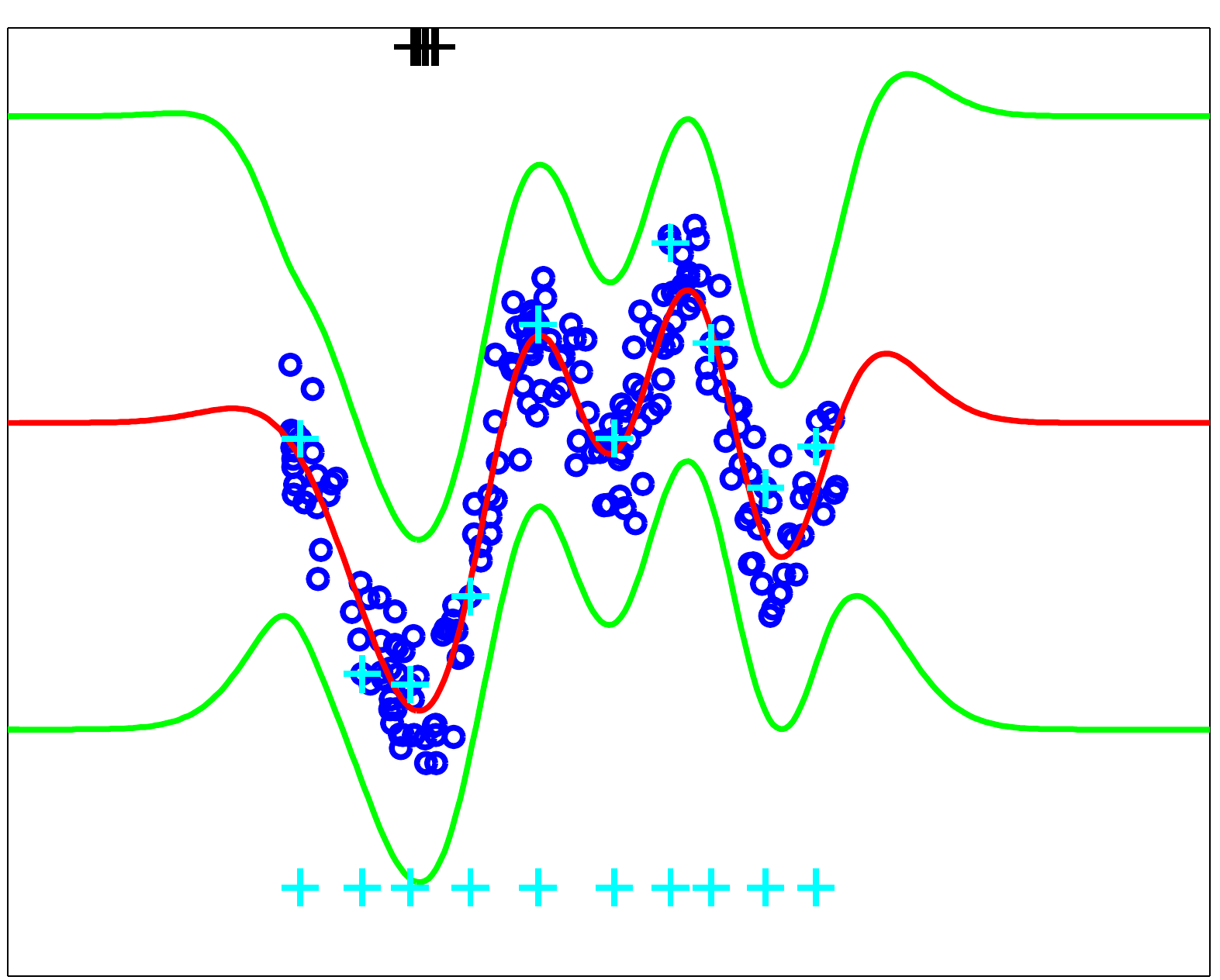}
\label{fig:cholqr_var_200}
}%
\subfigure[][CholQR-VAR]{
\includegraphics[width=0.18\textwidth]{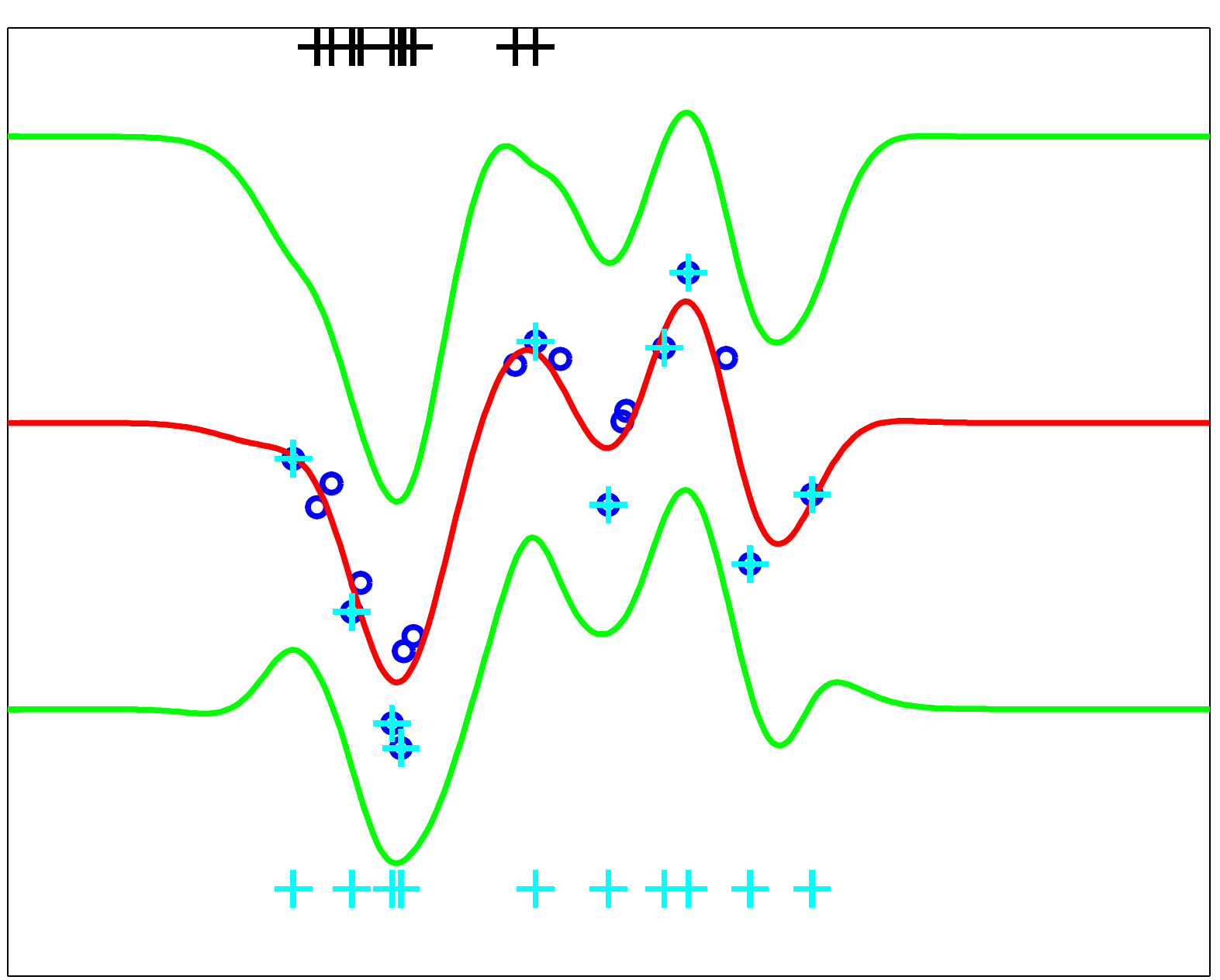}
\label{fig:cholqr_var_20}
}%
\subfigure[][SPGP]{
\includegraphics[width=0.18\textwidth]{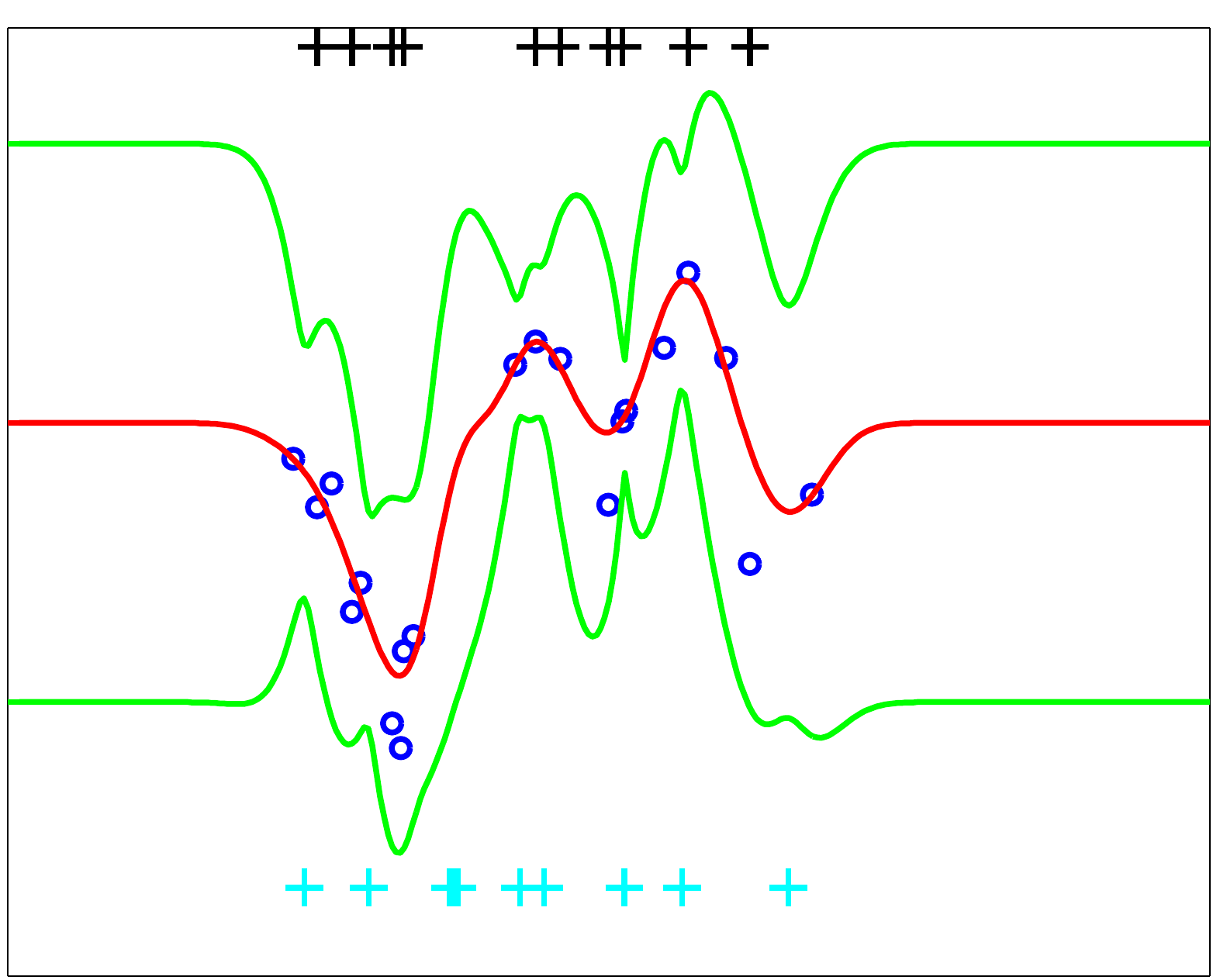}
\label{fig:fitc_20_rand_ini}
}%

\vspace*{-0.2cm}
\caption{Snelson's 1D example: prediction mean (red
  curves); one standard deviation in prediction uncertainty (green
  curves); inducing point initialization (black points at top of each
  figure); learned inducing point locations (the cyan points at the bottom, also overlaid on data for CholQR).}
\label{fig:snelson_1D}
\vspace*{-0.2cm}

\end{figure*}

Although CholQR was developed for discrete input domains, it can 
be competitive on continuous domains.  To that end, we compare to 
SPGP \cite{Snelson06} and IVM \cite{Lawrence03}, using RBF kernels with 
one length-scale parameter per input dimension; $\kappa(\vc{x}_i, \vc{x}_j) =
c\exp(-0.5\sum^{d}_{t=1}b_t({\vc{x}^{(t)}_i - \vc{x}^{(t)}_j})^2)$.
We show results from both the PP log likelihood and variational
objectives, suffixed by {\em MLE} and {\em VAR}.

We use the 1D toy dataset of \cite{Snelson06} to show how the PP 
likelihood with gradient-based optimization of inducing points is 
easily trapped in local minima.
Fig.\ \ref{fig:cholqr_mle_200} and \ref{fig:cholqr_var_200}
show that for this dataset our algorithm does not get trapped 
when initialization is poor (as in Fig.\ 1c of \cite{Snelson06}).
\comment{Unlike the problems SPGP which gets
stuck, our optimization on the same likelihood spreads out the inducing
points nicely.}
To simulate the sparsity of data in high-dimensional problems
we also down-sample the dataset to 20 points (every 10th point).
Here CholQR out-performs SPGP (see Fig.\,\ref{fig:cholqr_mle_20},
\ref{fig:cholqr_var_20}, and \ref{fig:fitc_20_clumped_ini}). 
By comparison, Fig.\,\ref{fig:fitc_20_rand_ini} shows SPGP learned with 
a more uniform initial distribution of inducing points avoids this 
local optima and achieves a better negative log likelihood of $11.34$ 
compared to $14.54$ in Fig.\,\ref{fig:fitc_20_clumped_ini}. 

\begin{figure*}%

\vspace*{-0.25cm}
\begin{center}
\subfigure{
\includegraphics[width=0.325\textwidth]{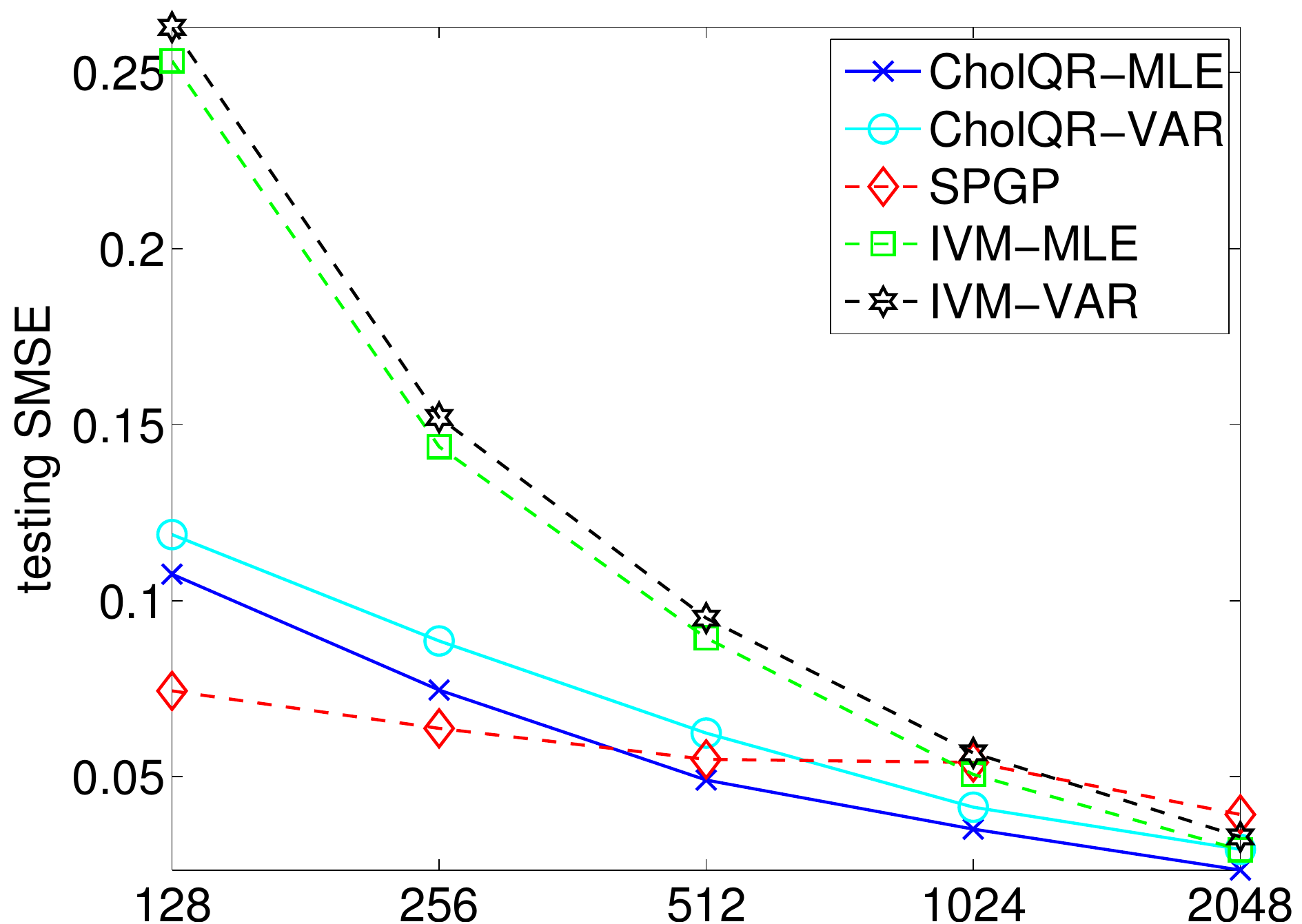}  
}%
\subfigure{
\includegraphics[width=0.325\textwidth]{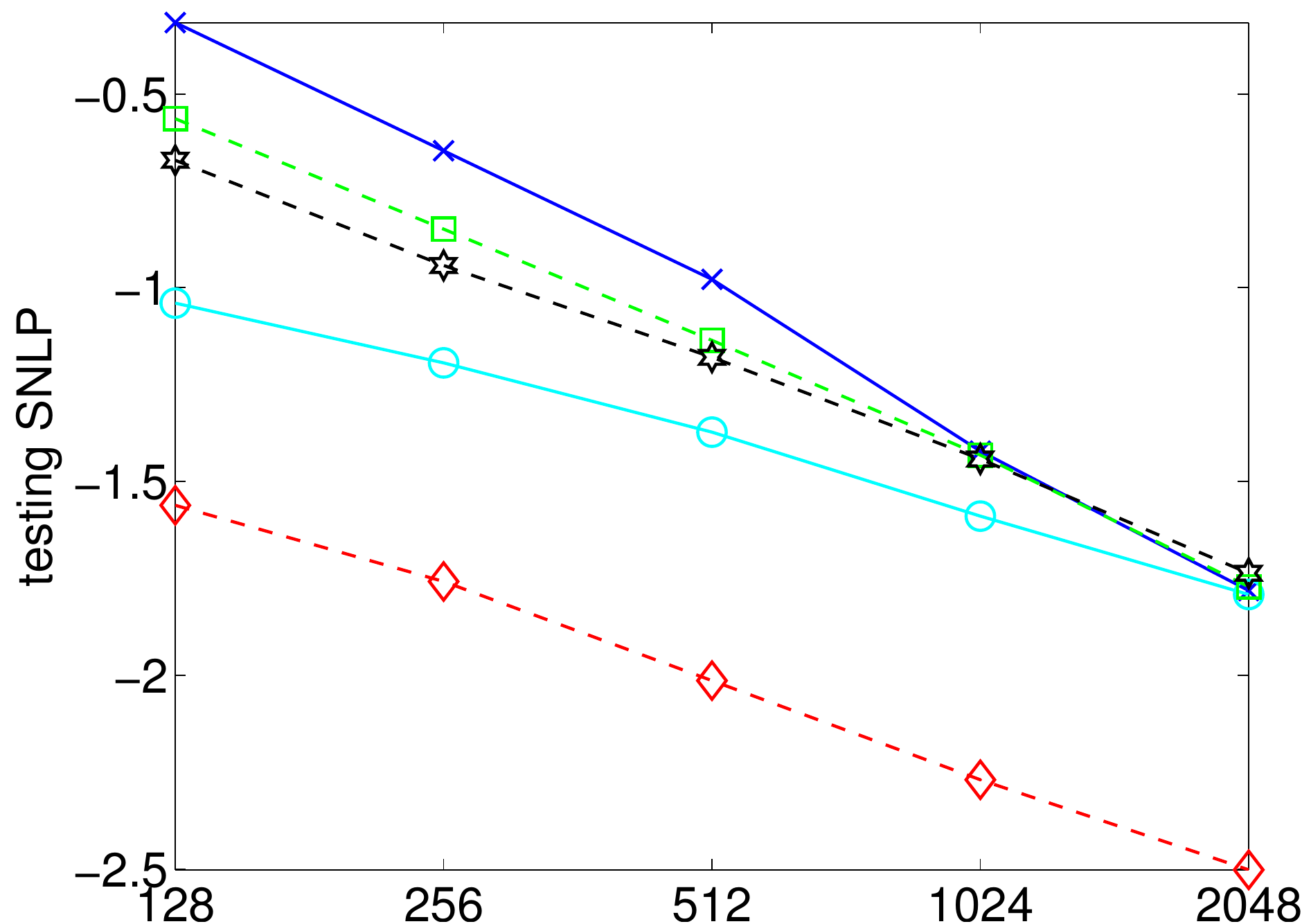}
}%
\end{center}

\vspace*{-0.25cm}
\caption{
Test scores on KIN40K as function of number of inducing points: for
each number of inducing points the value plotted is averaged over 10 runs
from 10 different (shared) initializations.
\label{fig:large_exps}
}

\vspace*{-0.2cm}
\end{figure*}

Finally, we compare CholQR to SPGP \cite{Snelson06} and 
IVM \cite{Lawrence03} on a large dataset.
{\em KIN40K} concerns nonlinear forward kinematic prediction.
It has 8D real-valued inputs and scalar outputs, with 10K training 
and 30K test points.
We perform linear de-trending and re-scaling as pre-processing.
For SPGP we use the implementation of \cite{Snelson06}.
Fig.\,\ref{fig:large_exps} shows that CholQR-VAR outperforms IVM 
in terms of SMSE and SNLP.  Both CholQR-VAR and CholQR-MLE outperform 
SPGP in terms of SMSE on KIN40K with large $m$, but SPGP exhibits
better SNLP.
This disparity between the SMSE and SNLP measures for CholQR-MLE
is consistent with findings about the PP likelihood in \cite{Titsias09}.

\vspace*{-0.2cm} 
\section{Conclusion}
\vspace*{-0.25cm} 

We describe  an algorithm for selecting inducing points for Gaussian 
Process sparsification.  It optimizes principled objective functions, 
and is applicable to discrete domains and non-differentiable kernels.
On such problems it is shown to be as good as or better than competing 
methods and, for methods whose predictive behavior is similar, our method 
is several orders of magnitude faster.  On continuous domains the method 
is competitive. Extension to the SPGP form of covariance approximation would be interesting future research.

\end{document}